%% file: main.tex
\documentclass[runningheads]{llncs}

\usepackage{eccv}

\usepackage{eccvabbrv}

\usepackage{booktabs}
\usepackage{multirow}
\usepackage{makecell}
\usepackage{graphicx}
\usepackage[table]{xcolor}
\usepackage{algorithm}
\usepackage{algpseudocode}
\usepackage{amssymb}
\usepackage{marvosym}
\usepackage{titletoc}

\usepackage{url}
\usepackage{capt-of}
\usepackage{multirow}
\usepackage{amsmath,amssymb}
\usepackage{threeparttable}
\usepackage{algorithm}
\usepackage{algpseudocode}

\input{preamble}

\usepackage{tikz}
\usepackage{adjustbox}
\usepackage{xcolor}
\usepackage[accsupp]{axessibility}  % Improves PDF readability for those with disabilities.

\newcommand{\appitem}[2]{%
  \noindent
  \hyperref[#2]{%
    \textbf{\ref{#2}}\hspace{0.6em}\textbf{#1}\dotfill\pageref{#2}%
  }\par%
}

\newcommand{\appsubitem}[2]{%
  \noindent
  \hyperref[#2]{%
    \hspace*{1.2em}\ref{#2}\hspace{0.6em}#1\dotfill\pageref{#2}%
  }\par%
}

\newcommand{\myparagraph}[1]{\noindent{\bf{#1}}~~}

\newcommand{\tok}[1]{\texttt{[#1]}}

\definecolor{yellow}{rgb}{1, 1, 0.7}
\definecolor{tableorange}{rgb}{1, 0.85, 0.7}
\definecolor{tablered}{rgb}{1, 0.7, 0.7}
\definecolor{red}{rgb}{1, 0, 0}

\usepackage{hyperref}

\definecolor{pearDark}{HTML}{2980B9}
\definecolor{pearThree}{HTML}{E74C3C}
\definecolor{pearDarker}{HTML}{1D2DEC}

\usepackage{colortbl}
\hypersetup{
  colorlinks,
  breaklinks=true,
  citecolor=pearDark,
  linkcolor=pearThree,
  urlcolor=pearDarker}

\usepackage{orcidlink}

\newcommand{\pmstd}[2]{%
  #1\kern0.05em{\raisebox{-0.00ex}{\tiny$\pm$#2}}%
}
\newcommand{\equalcontribmark}{\textsuperscript{*}}
\newcommand{\correspondingmark}{\textsuperscript{\Letter}}
\begin{document}

% ---------------------------------------------------------------
% TODO REVIEW: Replace with your title
% \title{Plug-and-Play Global Priors via Test-Time Registers}
\title{Test-Time Registers as Global Priors for Tokenized Image Generation}

% TODO REVIEW: If the paper title is too long for the running head, you can set
% an abbreviated paper title here. If not, comment out.
\titlerunning{RegToken: Registers as Global Priors for Image Generation}

% TODO FINAL: Replace with your author list.
% Include the authors' OCRID for the camera-ready version, if at all possible.
\author{Cheng-Yao~Hong\inst{1}\equalcontribmark  \and
Yifan~Wang\inst{1}\orcidlink{0009-0005-1567-2787}\equalcontribmark  \and
Yuewei~Lin\inst{2}\orcidlink{0000-0002-1429-4543} \and Chenyu~You\inst{1}\orcidlink{0000-0001-8365-7822}\correspondingmark}

% TODO FINAL: Replace with an abbreviated list of authors.
\authorrunning{C.~Hong et al.}
% First names are abbreviated in the running head.
% If there are more than two authors, 'et al.' is used.

% TODO FINAL: Replace with your institution list.
\institute{Stony Brook University \and
Brookhaven National Laboratory\\
\small{\url{https://y-research-sbu.github.io/RegToken}}}

\maketitle
\begingroup
\renewcommand{\thefootnote}{*}
\footnotetext[1]{Equal contribution.}
\renewcommand{\thefootnote}{\Letter}
\footnotetext[2]{Corresponding author: \email{chenyu.you@stonybrook.edu}.}
\endgroup

\begin{abstract}
Attention-based models often develop \textit{attention sinks}, where a small number of tokens repeatedly attract attention and accumulate unusually large activations. In vision transformers, these outliers are closely related to registers, which have been \emph{diagnostically} linked to global, low-frequency image structure. Existing work has largely studied registers through interpretability analyses and linear probes, leaving open whether they can be operationalized as plug-and-play signals for \emph{generation} without retraining. We revisit this question in tokenized image generation. Using OpenCLIP and DINOv2 on ImageNet, we find that test-time register features exhibit stronger low-frequency concentration than both \texttt{[CLS]} readouts and patch-mean features, and show a consistent (albeit moderate) correlation with pixel-space DCT low-frequency energy. Motivated by these diagnostics, we introduce \textbf{RegToken}, a training-free procedure that converts register structure into a small set of \textit{global prior tokens} by (i) NFN-based layer localization, (ii) TokenRank-guided subspace extraction, and (iii) a projection-and-conservation update on the register subspace. Inserted into a frozen compact 1D token generation pipeline, RegToken improves ImageNet generation and alignment metrics (\eg, FID-5k $20.5\!\rightarrow\!20.1$, SigLIP $3.6\!\rightarrow\!3.9$) without modifying pretrained weights, and accelerates test-time optimization (Steps@$ \tau $ $74\!\rightarrow\!52$). Overall, our results suggest that structures often viewed as attention artifacts can be repurposed as lightweight global priors for tokenized generation.
\end{abstract}

% \end{abstract}

\section{Introduction}
\label{sec:intro}

Attention-based architectures such as Vision Transformers (ViTs)~\cite{DosovitskiyB0WZ21}, large language models (LLMs)~\cite{BrownMRSKDNSSAA20,abs-2405-00732}, and multimodal transformers~\cite{RadfordKHRGASAM21,ChertiBWWIGSSJ23} rely on repeated token interactions through self-attention, yet they consistently exhibit \emph{attention sinks}, where a small subset of tokens attracts a disproportionate share of attention and accumulates unusually large activations. Across language and multimodal models, sink tokens (often near sequence boundaries or weakly semantic positions) can behave as persistent computational reservoirs~\cite{XiaoTCHL24,abs-2402-17762,KangK0H25,wang2025fourier,GuPDLZD0L25}; in vision transformers, analogous high-norm outlier tokens attract large attention mass and often align with visually uninformative regions~\cite{DarcetOMB24,abs-2506-08010}. Importantly, these outliers are not merely artifacts: evidence suggests they encode structured global information, including coarse layout and low-frequency scene statistics~\cite{DarcetOMB24}.

\begin{figure}[t]
\centering
\includegraphics[width=\textwidth]{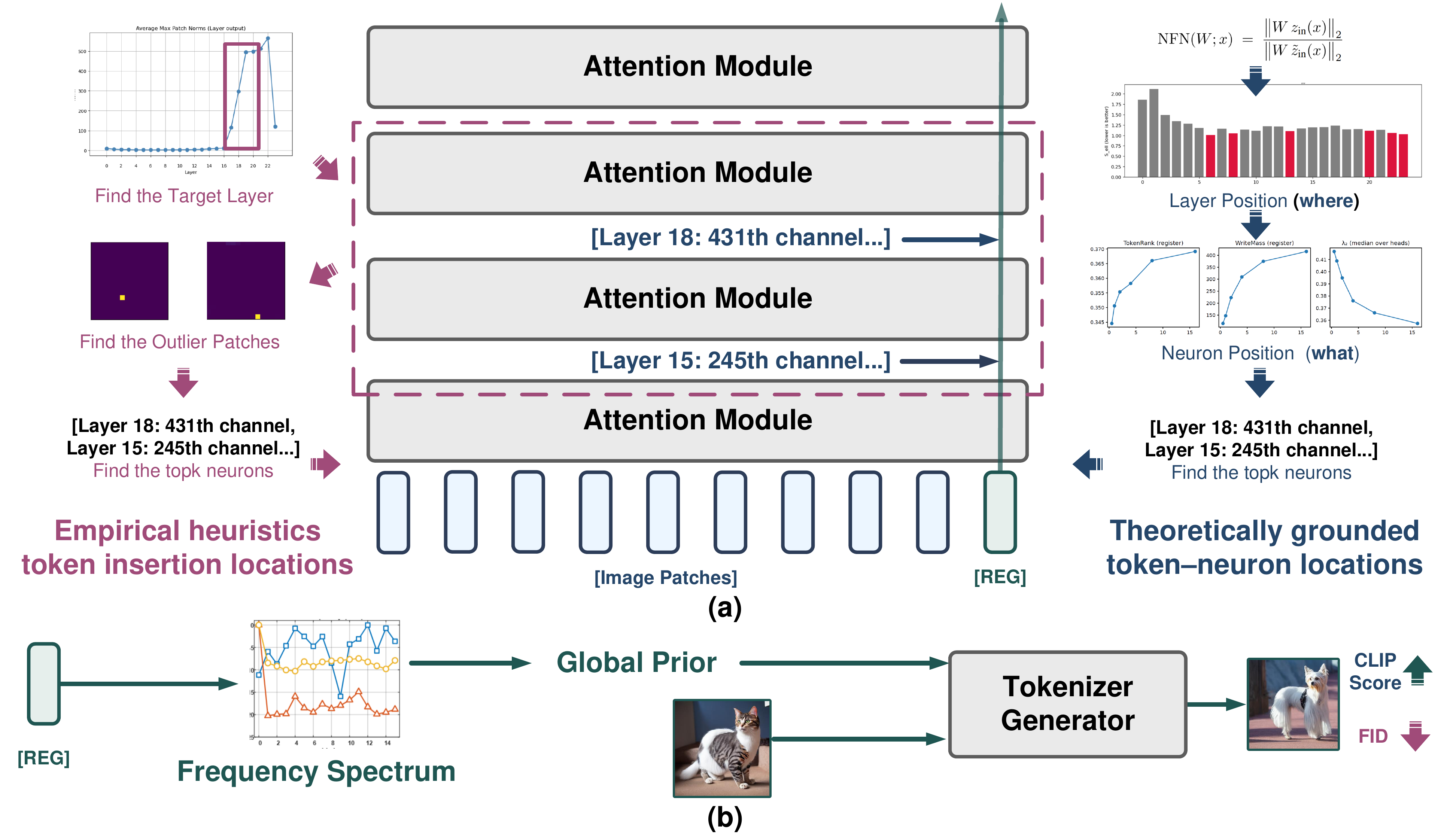}

\caption{\textbf{Overview of our proposed RegToken.}
(a) Prior test-time register insertion methods~\cite{abs-2506-08010} rely on heuristic layer and channel selection, which limits their transferability across architectures.
We instead localize register structure using NFN-based layer scores and TokenRank-guided channel importance, and construct register tokens through a projection-and-conservation interpolation rule.
(b) The resulting test-time registers act as compact global control tokens for a compact 1D token generation pipeline, providing global context that is consistent with low-frequency scene cues and improving generative quality and text--image alignment on ImageNet benchmarks.}
\label{fig: teaser}
\end{figure}

Several approaches have proposed to interpret or control this behavior. Darcet \etal~\cite{DarcetOMB24} introduce \emph{register tokens} during training, providing dedicated positions that absorb internal computation and stabilize dense representations. More recently, \cite{abs-2506-08010} show that a sparse set of \emph{register neurons} can drive the emergence of high-norm outliers and demonstrate that similar effects can be reproduced by injecting \emph{test-time registers} into frozen models. These perspectives suggest that attention sinks arise from an interaction between token-level outliers and neuron-level activation patterns.

Despite these insights, an important question remains open. If registers indeed encode structured global information, can they be reused as a compact signal for downstream tasks -- particularly for \emph{generation} -- without retraining large pretrained models? Existing work mainly studies registers through interpretability analysis or discriminative probes. Turning them into a practical generative prior requires identifying where register-related structure appears in a frozen backbone, isolating the relevant feature directions, and consolidating the signal into a small number of tokens that a generative decoder can consume.

In this work, we explore this question in the context of tokenized image generation. As a diagnostic, we analyze token representations from OpenCLIP and DINOv2 on ImageNet images. We observe that features associated with register tokens consistently concentrate more energy in low-frequency components than both \texttt{[CLS]} tokens and patch-mean features. These patterns suggest that register representations emphasize smooth, scene-level information that complements both discriminative readouts and local patch details.

Motivated by this observation, we propose \textbf{RegToken}, a training-free framework that converts register-associated structure into a small set of \emph{global prior tokens}. Our approach first identifies where register activity emerges in a frozen backbone using \emph{Normalized Feature Norms} (NFN), a forward-only criterion that highlights modules whose responses are dominated by a small number of feature directions. We then localize register-relevant channels using a TokenRank-guided importance measure that links neuron activations to token centrality in attention dynamics. Finally, we consolidate these signals into dedicated tokens through a \emph{projection-and-conservation} interpolation rule that transfers register-subspace energy from outlier patch tokens while approximately preserving global statistics.
When integrated into an HCT-style generation pipeline built on compact 1D
visual tokens~\cite{YuWDSCC24,abs-2506-08257}, the resulting tokens act as compact global control signals. Experiments on ImageNet benchmarks show consistent improvements in generative quality and text–image alignment, measured by FID, IS, CLIPScore, and SigLIP, while keeping both the backbone and the decoder fully frozen.
\begin{itemize}
\item We analyze register-associated features in large vision transformers and show that they emphasize low-frequency, scene-level statistics distinct from both \texttt{[CLS]} readouts and local patch tokens.
\item We introduce RegToken, a training-free method that extracts register structure from frozen ViTs and converts it into a compact set of reusable tokens through NFN-based layer localization and TokenRank-guided register subspaces.
\item We demonstrate that test-time registers can serve as global priors for tokenized image generation, improving FID, IS, CLIPScore, and SigLIP on ImageNet while keeping pretrained backbones and decoders frozen.
\end{itemize}

\section{Related Work}
\label{sec: related}

\myparagraph{Attention Sinks and Registers.}
Attention sinks have been observed across language, multimodal, and vision transformers~\cite{XiaoTCHL24,abs-2402-17762,GuPDLZD0L25,KangK0H25,luo2025sink}.
In language models, a small subset of tokens often attracts persistent attention and acts as a computational reservoir during long-context decoding~\cite{XiaoTCHL24,abs-2402-17762}.
In vision transformers, similar effects appear as high-norm outlier tokens that accumulate large fractions of attention mass and frequently correspond to visually uninformative regions~\cite{DarcetOMB24,abs-2506-08010}.
Darcet \etal~\cite{DarcetOMB24} introduce learned \emph{register tokens} during training to absorb internal computation and stabilize dense representations.
Complementarily, \cite{abs-2506-08010} trace these outliers to a sparse set of \emph{register neurons} and show that similar effects can be reproduced by injecting \emph{test-time registers} into frozen models.
These studies suggest that sink-like tokens may encode structured global information rather than being purely artifacts.

\myparagraph{Analyzing Transformer Attention.}
Recent work has explored broader tools for interpreting or reshaping transformer representations beyond raw attention weights, including attention analysis, sparse representations, robustness diagnostics, and module-level feature statistics~\cite{you2024calibrating,you2025uncovering,wen2025beyond,guo2026csrv2}. \cite{yang2025less} propose a sparse-attention rule for long-form reasoning that aggregates per-head token selections into a global ranking used by later layers.
At the module level, \cite{abs-2506-20629} introduce Normalized Feature Norms (NFN), a forward-only criterion that measures module–data alignment and helps identify layers suitable for parameter-efficient adaptation.
Orthogonally, \cite{abs-2507-17657} interpret attention matrices as Markov chains and define TokenRank as the stationary distribution capturing multi-hop token influence.
Together these perspectives provide complementary diagnostics of where important computation occurs and which tokens dominate attention flow.
Our approach builds on these insights to localize register structure in frozen backbones.

\myparagraph{Tokenized Image Generation.}
Recent work has explored compact or structured representations for generation, reconstruction, and multimodal understanding across images and related visual domains~\cite{liu2021auto,liu2021aligning,liu2022retrieve,huang2024cross,han2024hybrid,wang2026let,sun2026coma,gao2026neurosonic}.
For example, TiTok~\cite{YuWDSCC24} introduced compact one-dimensional visual tokens for image reconstruction and generation, showing that images can be represented by a small number of discrete tokens. Following this line, HCT~\cite{abs-2506-08257} uses highly compressed token sequences together with test-time token optimization for image generation. More recent text-aware tokenizers, such as TexTok~\cite{ZhaYFRSKG25} and TA-TiTok~\cite{KimHYYSKC25}, condition image tokenization on language, allowing
visual tokens to focus on image details while text provides high-level semantics. AToken~\cite{lu2025atoken} further studies unified tokenization across images, videos, and 3D assets.
Frequency-based compression methods also show that vision features concentrate energy in low frequencies and can be aggressively compressed with limited performance loss~\cite{wang2025fourier}.
Orthogonal to these broader representation and tokenizer-design efforts, our work does not train a new tokenizer or condition the tokenizer on text. Instead, RegToken extracts a training-free global prior from frozen ViT register structure and injects it into a compact 1D token generation pipeline.

\section{Motivation and Challenges}
\label{sec: obs}
Recent studies show that attention sinks in transformer architectures are not merely artifacts but often reflect structured internal computation.
In vision transformers, high-norm outlier tokens tend to attract large fractions of attention mass, and introducing explicit \emph{register tokens} during training has been shown to absorb these outliers and stabilize dense features~\cite{DarcetOMB24}.
Complementarily, \cite{abs-2506-08010} demonstrate that a sparse set of \emph{register neurons} can drive the emergence of such outliers and show that similar effects can be reproduced through \emph{test-time registers} in frozen models.
Related phenomena have also been reported in language and multimodal transformers, where a small subset of tokens acts as persistent attention sinks that influence information routing in long contexts~\cite{XiaoTCHL24,GuPDLZD0L25,KangK0H25}.
Sink- or register-associated representations are not necessarily unstructured noise.
For example, \cite{abs-2506-08010} report that test-time register features achieve competitive linear-probe accuracy on ImageNet relative to the \texttt{[CLS]} token (\Cref{tab:fft_stats}), suggesting that they encode systematic information.

This raises a natural question: \emph{what information do register representations capture, and how do they differ from other tokens?}

\myparagraph{Notation and Preliminaries.}
Let $X^\ell \in \mathbb{R}^{N \times d}$ denote the token embeddings entering attention layer $\ell$, where $N = 1 + R + P$ consists of one \tok{CLS}, $R$ \texttt{[REG]}, and $P$ \tok{PATCH} tokens.
Self-attention is computed as:
\begin{gather}
Q^\ell=X^\ell W_Q^\ell, \quad
K^\ell=X^\ell W_K^\ell, \quad
V^\ell=X^\ell W_V^\ell, \\
A^\ell=\mathrm{softmax}\!\left(\tfrac{Q^\ell {K^\ell}^\top}{\sqrt{d_k}}\right), \quad
\text{AttnOut}=A^\ell V^\ell,
\label{eqn:condition}
\end{gather}
where $W_Q^\ell$, $W_K^\ell$, and $W_V^\ell$ are learned projection matrices and $A^\ell$ denotes the attention matrix for a single head.
A token is referred to as an \emph{attention sink} if it consistently receives a large share of incoming attention across layers.
In vision models, a \emph{register token} is an explicit position introduced to absorb global computation~\cite{DarcetOMB24}, while a \emph{register neuron} denotes a sparse subset of channels whose large activations concentrate on outlier tokens~\cite{abs-2506-08010}.
A \emph{test-time register} refers to a register-like token or high-norm outlier behavior identified in a frozen model without retraining.
In this paper we focus on large vision backbones. We use \emph{RegToken} to denote our extracted global prior token. It is not a learned register token; rather, it is a training-free token
constructed from frozen ViT register structure and mapped to a compact-token generation pipeline. 

\begin{table}[t]
\centering
\small
\setlength{\tabcolsep}{2.5pt} % Squeeze table slightly
\renewcommand{\arraystretch}{0.85}
\caption{\textbf{Low-frequency concentration of token features.}
Statistics are computed on 1000 ImageNet images using the same token features as in Figure~\ref{fig: fft}.
We report Mean$\pm$Std and paired $t$-test $p$-values.
Top: DINOv2. Bottom: OpenCLIP.}

\label{tab:fft_stats}
\resizebox{\textwidth}{!}{
\begin{tabular}{lccccc}
\toprule
Metric & \texttt{[REG]} & Patch-mean & \texttt{[CLS]} &
$p$ (\texttt{[REG]} vs Patch) & $p$ (\texttt{[REG]} vs \texttt{[CLS]}) \\
\midrule
$\rho_{1:15}\uparrow$
& \cellcolor{pearDark!20}$\mathbf{0.507\pm0.076}$
& $0.443\pm0.073$
& $0.504\pm0.078$
& $1.13\times10^{-68}$
& $7.32\times10^{-4}$ \\
$\bar dB_{\text{low}}\uparrow$
& \cellcolor{pearDark!20}$\mathbf{-10.03\pm2.63}$
& $-18.60\pm1.95$
& $-10.48\pm2.28$
& $<10^{-300}$
& $4.8\times10^{-16}$ \\
\midrule
$\rho_{1:15}\uparrow$
& \cellcolor{pearDark!20}$\mathbf{0.524\pm0.017}$
& $0.508\pm0.067$
& $0.495\pm0.089$
& $1.60\times10^{-13}$
& $1.99\times10^{-22}$ \\
$\bar dB_{\text{low}}\uparrow$
& \cellcolor{pearDark!20}$\mathbf{-3.02\pm0.38}$
& $-7.24\pm1.26$
& $-11.33\pm1.52$
& $<10^{-300}$
& $<10^{-300}$ \\
\bottomrule
\end{tabular}
}
\end{table}

\subsection{What Registers Store?}
\label{sec:charact}

\myparagraph{Prior Evidence.}
Previous studies report that high-norm outlier tokens frequently appear in low-information regions and behave as internal workspaces within ViTs~\cite{DarcetOMB24}.
Introducing learned register tokens during training absorbs these outliers and produces smoother features and attention maps.
A training-free variant identifies sparse register neurons that drive high-norm activations and shows that test-time registers reproduce similar effects across OpenCLIP and DINOv2 backbones~\cite{abs-2506-08010}.
More recently, DINOv3~\cite{simeoni2025dinov3} incorporates learned registers and Gram anchoring to stabilize dense features, suggesting that modern ViT architectures treat registers as dedicated global workspaces.

\begin{figure}[t]
\centering
\includegraphics[width=\linewidth]{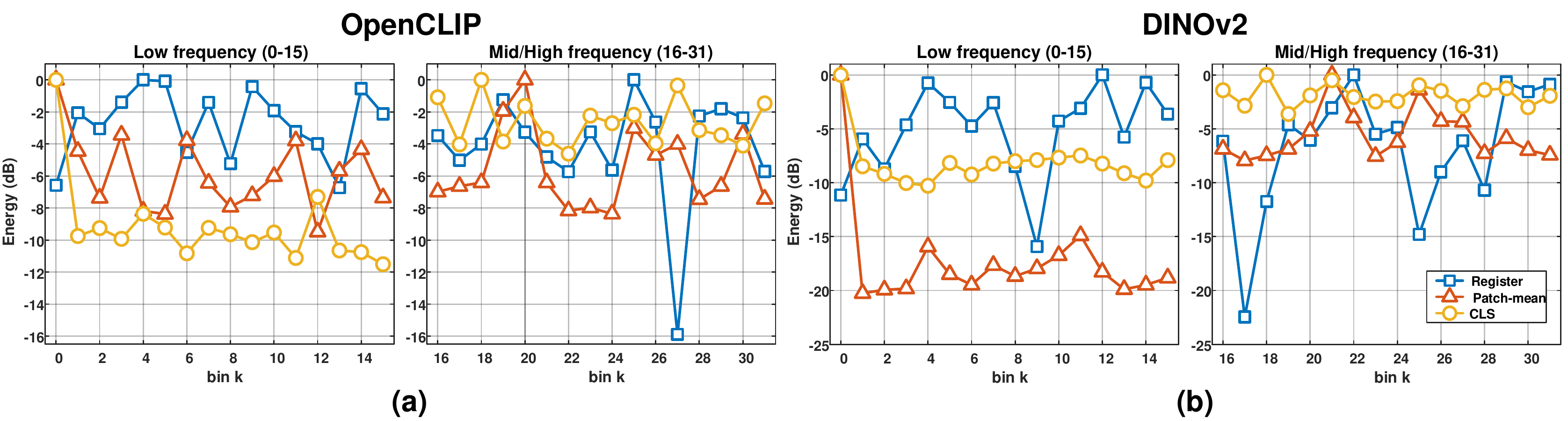}

\caption{\textbf{Spectral structure of token embeddings.}
We visualize PCA-projected token features and their 1-D FFT spectra along the feature dimension for three token types: register, patch-mean, and \texttt{[CLS]}.
Across both OpenCLIP and DINOv2, \texttt{[REG]} features exhibit stronger low-frequency concentration than patch-mean features and differ systematically from \texttt{[CLS]} representations.
This pattern suggests that register tokens encode smooth, scene-level statistics, whereas \texttt{[CLS]} primarily reflects discriminative readout and patch tokens encode finer-grained details.}

\label{fig: fft}

\end{figure}

\myparagraph{Hypothesis and Analysis.}
Motivated by these findings, we examine what type of information register-associated features capture.
Our hypothesis is that register representations emphasize \emph{smooth, scene-level statistics} such as color tone, illumination, and coarse spatial layout, rather than local high-frequency details.

To test this hypothesis, we analyze token features from OpenCLIP and DINOv2 on 1000 ImageNet images.
For each image we compute PCA-projected embeddings and analyze their spectral properties by applying a one-dimensional FFT over the \emph{feature dimension}.
This diagnostic measures embedding smoothness rather than spatial image frequency.
We compare three token types: test-time \texttt{[REG]}, patch-mean features, and \texttt{[CLS]}.

To quantify low-frequency concentration, we compute two scalar metrics per image: $\rho_{1:15}$, the low-band energy ratio (excluding DC), and $\bar{d}B_{\text{low}}$, a measure of low-band spectral flatness.
Table~\ref{tab:fft_stats} reports the mean and standard deviation across images together with paired $t$-test $p$-values.
Across both backbones, \texttt{[REG]} tokens consistently exhibit stronger low-frequency concentration than patch-mean features and \texttt{[CLS]} tokens, often with $p < 10^{-16}$.
The trend remains after whitening; Figure~\ref{fig: fft} shows representative spectra and PCA visualizations.

To relate this embedding-spectrum diagnostic to pixel-space statistics, we further correlate $\rho_{1:15}$ and $\bar{d}B_{\text{low}}$ with an image-level low-frequency energy ratio computed using a 2D DCT on RGB pixels.
We observe a consistent positive correlation (Table~\ref{tab:bridge_dct_corr} and
Appendix~\ref{secA:bridge_dct} in the supplementary material), suggesting that register features are associated with low-frequency image structure.

Taken together, these results suggest that register tokens encode compact, slowly varying information distinct from both \texttt{[CLS]} (task readout) and local patch features.
This distinction is particularly relevant for tokenized image generation, where only a small number of tokens must represent global structure during test-time optimization.
In HCT-style 1D decoding, a compact global signal can therefore guide generation without modifying pretrained weights.
These observations motivate the use of test-time registers as compact global prior tokens in the generation experiments that follow.

\section{Method}
\label{sec: method}

The observations in Section~\ref{sec: obs} suggest that register-associated features encode smooth, scene-level statistics that summarize global image structure.
We leverage this property by turning test-time registers into a small set of global prior tokens that guide token-based generation while keeping the backbone and decoder fully frozen.
Unlike prior register analyses that focus on representation smoothing~\cite{DarcetOMB24,abs-2506-08010}, our goal is to convert register structure into reusable tokens for generative decoding.
Our method first localizes register-related structure in a frozen ViT backbone, then constructs register tokens through token--neuron interpolation, and finally injects these tokens into a tokenized generation pipeline.
The key technical challenge is to identify the register-neuron subspace in a model-agnostic and training-free manner.

\myparagraph{Overview and Generation Protocol.}
Our RegToken has two components.  First, we localize a register subspace in a frozen ViT backbone (Sections~\ref{sec:token2neurons} -- \ref{sec:interp}).
Second, we map the resulting register tokens to a frozen tokenizer/decoder for generation (Section~\ref{sec:register4generalize}).
To avoid ambiguity in evaluation, we distinguish \emph{token construction and insertion} from \emph{optional test-time optimization}.
Unless otherwise stated, we consider two regimes with frozen weights: \emph{prior injection} (all $\mathbf{z}_{0:T}$ fixed) and \emph{prior + opt} (optimize only the inserted $z_0$ while keeping $\mathbf{z}_{1:T}$ fixed).
Implementation details and hyperparameters are provided in Appendix~\ref{app:generation_details} in the supplementary material.

\subsection{Register Subspace Localization}
\label{sec:token2neurons}
The original test-time register procedure of \cite{abs-2506-08010} identifies layers where token norms or attention mass increase sharply, selects high-norm patch tokens around those layers, and ranks channels by their average activations at these outlier locations. While effective, this heuristic depends on architecture-specific hyperparameters and manually chosen layer windows.

\myparagraph{NFN-based Layer/Module Selection.}
Following \cite{abs-2506-20629}, we first determine where to intervene by measuring \emph{Normalized Feature Norms} (NFN) across layers and submodules.
For a module $W$ (\eg, a per-head $Q$, $K$, $V$ projection, attention output, or MLP up/down projection) with input feature $z_{\text{in}}(x)$ for image $x$, define:
\begin{equation}
\mathrm{NFN}(W;x) =
\frac{\| W z_{\text{in}}(x)\|_2}{\| W \tilde z_{\text{in}}(x)\|_2},
\label{eq:nfn_instance}
\end{equation}
where $\tilde z_{\text{in}}(x)$ is a random vector with the same dimension and norm as $z_{\text{in}}(x)$ and i.i.d.\ zero-mean Gaussian coordinates.
Aggregating over a dataset $\mathcal{D}$ and attention heads yields the layer-module score:
\begin{equation}
\mathrm{NFN}_{\ell,m} =
\mathbb{E}_{x\sim\mathcal{D}}\!\left[\mathrm{median}_{h}\;
\mathrm{NFN}(W^{(\ell,m,h)};x)\right],
\end{equation}
with $m\in\{\texttt{Q},\texttt{K},\texttt{V},\texttt{AttnOut},
\texttt{MLP}_{\uparrow},\texttt{MLP}_{\downarrow}\}$.
We formulate per-layer evidence by:
\begin{equation}
S_\ell = \max_m \mathrm{NFN}_{\ell,m}.
\end{equation}
Values $S_\ell\approx1$ indicate responses comparable to a random baseline, while large $S_\ell$ highlight modules whose outputs are dominated by a small subset of input directions.
In practice these peaks coincide with layers where sink/register behavior becomes prominent.
We therefore select the $\kappa$ layers with the largest $S_\ell$ as candidate layers $\mathcal{L}_{\mathrm{cand}}$ (typically $\kappa\in[3,5]$).
Unlike PLoP~\cite{abs-2506-20629}, which uses low-NFN modules to place LoRA, we treat high-NFN layers as indicators of strong feature amplification associated with sink effects.

\myparagraph{Token-level Localization.}
For each $\ell\in\mathcal{L}_{\mathrm{cand}}$ we compute token $\ell_2$-norm maps over a window of $w$ layers around $\ell$ (\eg, $w=3$).
Outlier tokens are detected using a robust threshold based on the median and interquartile range.
We retain the top-$n$ spatial indices $\Omega_\ell=\mathrm{top}\text{-}n(\texttt{PATCH})$ that consistently appear as high-norm tokens within this window.

\myparagraph{Neuron-level Extraction.}
Given $\Omega_\ell$, we rank channels by their mean absolute activations over these outlier tokens in the preceding $\Delta$ layers:
\begin{equation}
r_d =
\frac{1}{\Delta |\Omega_\ell|}
\sum_{j=1}^{\Delta}\sum_{p\in\Omega_\ell}
|a^{(\ell-j)}_{p,d}|,
\end{equation}
where $a^{(\ell-j)}_{p,d}$ denotes the activation of channel $d$ at token $p$ and layer $\ell-j$.
The top-$k$ channels $\mathcal{R}_\ell=\mathrm{top}\text{-}k(\{r_d\})$ are taken as \emph{register neurons}, defining the register subspace $\mathrm{span}\{e_d: d\in\mathcal{R}_\ell\}$.

\myparagraph{From Subspace to Test-time Registers.}
Given $\mathcal{R}_\ell$, we introduce auxiliary register tokens that absorb activations along the register subspace.
Let $t_{\mathrm{reg}}$ denote the new token.
Before the residual update, we transfer activations on $\mathcal{R}_\ell$ from outlier patches to $t_{\mathrm{reg}}$, effectively concentrating global statistics in a small number of tokens while suppressing artifact-inducing outliers.
No model parameters are modified.

\myparagraph{Defaults.}
Unless otherwise stated we use $\kappa=5$, $n\in[3,8]$, $k\in\{32,64\}$, $w=3$, and $\Delta=2$.
NFN is computed on a 1000-image subset of ImageNet-\texttt{val}.
All backbone parameters remain frozen.

\subsection{Token-side Importance via Value TokenRank}
\label{sec:valuetokenrank}

Attention weights alone do not fully reflect the information carried by a token.
Following \cite{abs-2507-17657}, we combine attention with value magnitudes to estimate how strongly a token contributes to the computation.

\myparagraph{Write Mass.}
For an attention head with matrix $A\in\mathbb{R}^{T\times T}$ and value vectors $\{V_t\}_{t=1}^T$, we define the write mass of token $t$ as:
\begin{equation}
\mathrm{WriteMass}(t)=\sum_{i=1}^{T} A_{i,t}\|V_t\|_2^2.
\end{equation}

\myparagraph{TokenRank.}
Viewing $A$ as a Markov transition matrix, TokenRank $\pi$ is the stationary distribution satisfying $\pi^\top=\pi^\top A$.
Tokens with larger $\pi_t$ have greater influence under multi-hop attention flow.
In practice we compute $\pi$ per head and average across heads as in~\cite{abs-2507-17657}.

\myparagraph{Head Mixing Rate.}
Let $\lambda_2$ denote the second-largest eigenvalue of $A$.
Larger $\lambda_2$ corresponds to slower mixing and stronger persistence of attention mass.
Empirically we find that layers with large $\lambda_2$ often coincide with the emergence of high-norm outliers.

\subsection{Token-Neuron Interpolation}
\label{sec:interp}

Let $\mathcal{R}_\ell$ be the register neurons identified at layer $\ell$, and let $\mathcal{P}$ denote the set of outlier patch tokens.
We introduce a register token $t_{\mathrm{reg}}$ and reallocate activations along the register subspace to this token while approximately preserving statistics.

\myparagraph{Projection and Conservation.}
Let $P_{\mathcal{R}}$ be the orthogonal projector onto the register subspace.
Define the mean register activation:
\begin{equation}
    \bar v_{\mathcal{R}}=\frac{1}{|\mathcal{P}|}\sum_{p\in\mathcal{P}}P_{\mathcal{R}}V_p.
\end{equation}
% \]
For scale $s>0$ we update:
\begin{equation}
\label{eq:interp_update}
\begin{aligned}
V_{t_{\mathrm{reg}}} &\leftarrow V_{t_{\mathrm{reg}}}+s\,\bar v_{\mathcal{R}},\\
V_p &\leftarrow V_p-\alpha P_{\mathcal{R}}V_p,\quad p\in\mathcal{P}.
\end{aligned}
\end{equation}
We set $\alpha$ to approximately preserve the \emph{mean projection magnitude on the register subspace} across the affected tokens (rather than general second-order statistics).

\subsection{Registers for Tokenized Generation}
\label{sec:register4generalize}

We treat the resulting register tokens as global priors for compact 1D visual token pipelines such as TiTok and HCT~\cite{YuWDSCC24,abs-2506-08257}. Each register token is mapped to the nearest codebook entry and inserted into the tokenizer sequence.
Given a codebook $\mathcal{C}=\{c_j\}$, we select
\begin{equation}
    j_\ell^\star=\arg\max_j \sigma(r^{(\ell)},c_j),
\end{equation}
% \]
where $\sigma$ denotes cosine similarity.
Multiple registers can be fused by a convex combination weighted by TokenRank.
The fused register vector is inserted into the token sequence,
\begin{equation}
z_0\leftarrow(1-\gamma)z_0+\gamma\tilde r,
\end{equation}
with strength $\gamma\in[0,1]$.
Optionally, only the inserted token is optimized for a few steps under a frozen decoder with a CLIP-based objective (Appendix~\ref{subsecA:prior} in the supplementary material).

\begin{table}[t!]
    \centering
    \caption{\textbf{Norm separation between patch and register tokens.}
We compare the vanilla test-time register method~\cite{abs-2506-08010} with NFN-guided variants using different numbers of register neurons on OpenCLIP and DINOv2.}

    \label{tab: nfn}
    \resizebox{\textwidth}{!}{
    \begin{tabular}{lcccc}
    \toprule[1.0pt]
        \multicolumn{1}{l}{\multirow{2}{*}{\textbf{Method}}} &  \multicolumn{2}{c}{\textbf{OpenCLIP}~\cite{ChertiBWWIGSSJ23}} & \multicolumn{2}{c}{\textbf{DINOv2}~\cite{OquabDMVSKFHMEA24}} \\ \cmidrule(lr) {2-3} \cmidrule(lr) {4-5} & norm gap & norm gap (Attention) & norm gap & norm gap (Attention) \\
        \midrule

        Vanilla(50 neurons)~\cite{abs-2506-08010}  & \cellcolor{yellow}{62.03}& \cellcolor{tablered}\textbf{0.58}&468.83&\cellcolor{yellow}{0.60} \\
        NFN-guided (8 neurons) & 61.36& \cellcolor{yellow}{0.52}&\cellcolor{yellow}{485.23}&0.59 \\
        NFN-guided (16 neurons) & \cellcolor{tableorange}{64.42}& \cellcolor{tableorange}{0.55}&\cellcolor{tableorange}494.63&\cellcolor{tableorange}{0.63} \\
        NFN-guided (50 neurons) & \cellcolor{tablered}\textbf{70.09}& \cellcolor{tablered}\textbf{0.58}&\cellcolor{tablered}\textbf{633.71}&\cellcolor{tablered}\textbf{0.66} \\
    \bottomrule[1.1pt]
    \end{tabular}
    }
\end{table}

\begin{figure*}[t]
\centering
\includegraphics[width=\textwidth]{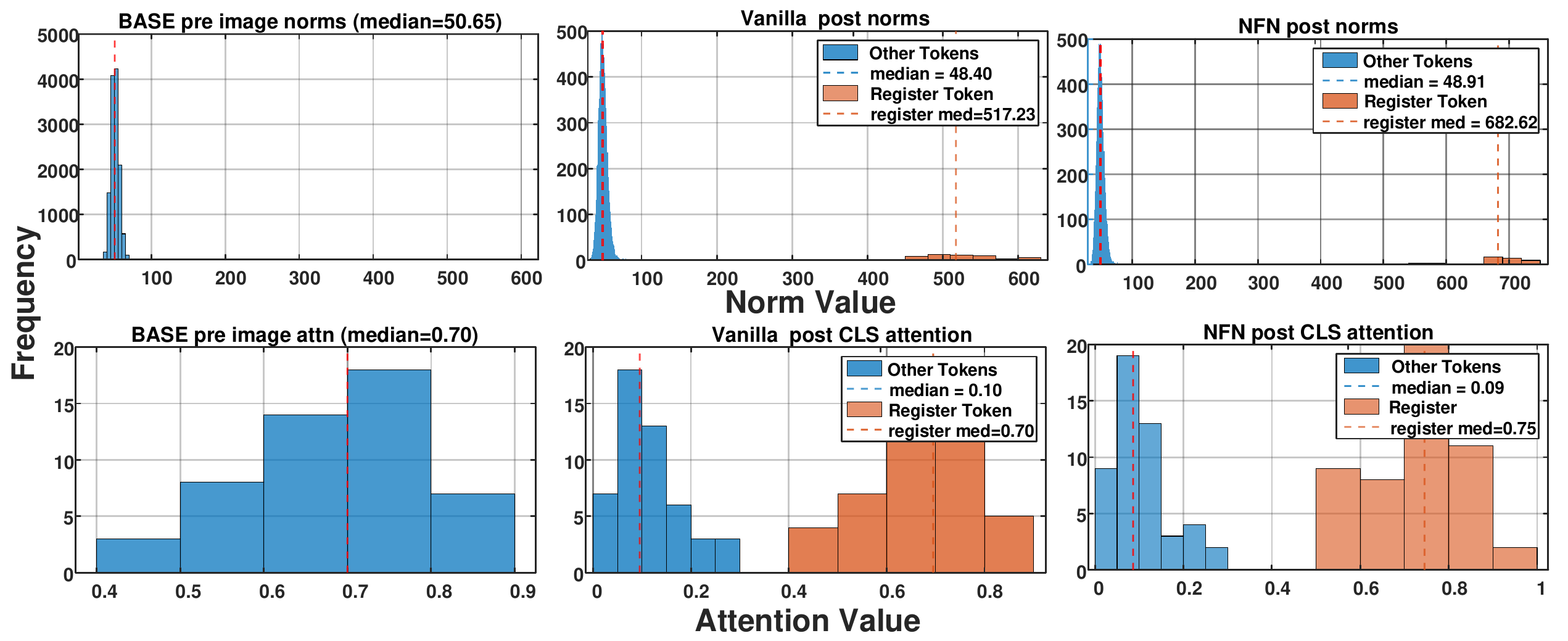}

\caption{\textbf{Effect of register insertion on DINOv2 features.}
A single register token is inserted at NFN-selected layers using the top-50 register neurons.
The register token accumulates large norm and attention mass, while the statistics of image tokens remain largely unchanged.}
\label{fig: nfn_dino}
\end{figure*}

%-------------------------------------------------
\section{Experiments}
\label{sec: exp}

\myparagraph{Evaluation Protocol.}
Unless otherwise stated, both the ViT backbone and the HCT decoder remain frozen.
When token optimization is enabled, we update only the injected global token while keeping all other decoder tokens fixed.
Table~\ref{tab: hct} compares methods under identical decoding hyperparameters; the only difference is the source of the injected prior.
Full protocol details, including optimization variables and ranges, are provided in Appendix~\ref{app:generation_details} in the supplementary material.
\myparagraph{Backbone usage.}
In all generation experiments, the HCT-style decoder and its codebook are fixed.
The backbone (DINOv2-L/14 or OpenCLIP-L/14) is used \emph{only} to compute the injected global prior token (CLS / TTR / RegToken) from the same input image, which is then mapped to the nearest VQ-LL-32 code via the same alignment procedure.
For \emph{No prior}, we set $\gamma{=}0$ and keep the initial token sequence unchanged.
All other decoding and optimization hyperparameters (seed set, steps, loss, and learning rate) are identical across backbones.

\subsection{NFN-based Layer Localization}

\myparagraph{From Outliers to Register Neurons.}
We first detect the onset of outliers $L_{\text{start}}$ from the patch-norm curve and identify outlier tokens within a small backward window.
Within this window, channels are ranked to select the top 50 register neurons, with a per-layer cap and back-filling to ensure exactly 50 positions.
This preserves the original outlier-to-neuron pipeline used by the test-time register method and provides a common backbone for both the baseline and our NFN-guided variant.
Figure~\ref{fig: nfn_dino} illustrates the effect of inserting a register token.
After insertion, both the register value norm and the CLS$\rightarrow$register attention increase substantially, while statistics of image tokens remain largely unchanged.
On DINOv2-L/14, the median register value norm increases from 517.2 to 682.6 (+32\%), and CLS$\rightarrow$register attention rises from 0.70 to 0.75 (+0.05).
In contrast, the median norm and attention of image tokens remain close to 49 and 0.10, respectively.
OpenCLIP exhibits a similar trend, with details provided in Appendix~\ref{secA:NFN_layer} in the supplementary material.
Table~\ref{tab: nfn} further shows that the NFN-guided variant produces comparable or stronger norm and attention separation than the standard test-time register approach, even with fewer neuron positions.

\begin{table}[t!]
\centering
\caption{\textbf{Effect on outliers and attention mixing (ImageNet-\texttt{val}, 1k images).}
Outlier fraction uses the pre-95th percentile threshold fixed for the post configuration.
$\lambda_2$ denotes the second eigenvalue of the attention Markov chain (median over heads).
We report the spectral gap $1-\lambda_2$, where a smaller gap indicates slower mixing.}

\label{tab: token}
\resizebox{\textwidth}{!}{
\begin{tabular}{lcccc}
\toprule

\multicolumn{1}{l}{\multirow{2}{*}{\textbf{Method}}} &  \multicolumn{2}{c}{\textbf{OpenCLIP}~\cite{ChertiBWWIGSSJ23}} & \multicolumn{2}{c}{\textbf{DINOv2}~\cite{OquabDMVSKFHMEA24}} \\ \cmidrule(lr) {2-3} \cmidrule(lr) {4-5} & $\Delta$ median(outlier) ($\downarrow$) & $\Delta$ gap $= -\Delta\lambda_2$ ($\downarrow$) & $\Delta$ median(outlier) ($\downarrow$) & $\Delta$ gap $= -\Delta\lambda_2$ ($\downarrow$)  \\

\midrule
Vanilla~\cite{abs-2506-08010} &$-0.013$& $-0.0024$& $-0.0215$  & $-0.0036$  \\
RegToken           &$-0.014$& $-0.0027$& $-0.0312$ & $-0.0946$ \\
\bottomrule
\end{tabular} %
} %
\end{table}

\begin{figure*}[t!]
\centering
\includegraphics[width=\textwidth]{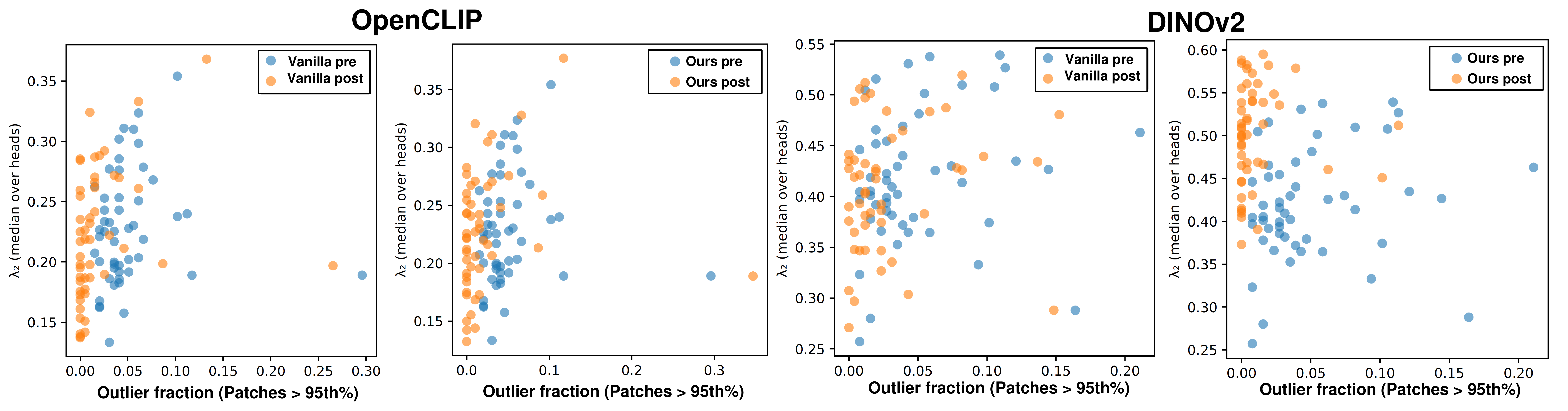}

\caption{\textbf{Correlation between $\lambda_2$ and outlier tokens.}
Each dot corresponds to one ImageNet-\texttt{val} image.
The $x$-axis shows the fraction of patch tokens above the \emph{pre}-95th norm threshold, and the $y$-axis shows $\lambda_2$ of the attention Markov chain (median over heads).
Vanilla test-time registers produce a modest shift toward fewer outliers and slightly lower $\lambda_2$, whereas the NFN-guided head-gated variant yields a larger reduction in outliers and a clearer drop in $\lambda_2$, indicating stronger mixing.}

\label{fig: token}
\end{figure*}

\myparagraph{Value$\times$TokenRank Bridge.}
Attention probabilities alone can be misleading: a token may attract attention but carry little information.
We therefore combine WriteMass with TokenRank to measure both content flow and global centrality (Section~\ref{sec:valuetokenrank}).
Scaling the injected register token increases both TokenRank and WriteMass while reducing $\lambda_2$, indicating that information is absorbed by the register rather than amplified across patches.
This behavior is illustrated in Figure~\ref{fig: token}.
Table~\ref{tab: token} reports the effect on outlier frequency and mixing.
On DINOv2-L/14 (1k validation images), vanilla test-time registers reduce the median outlier fraction by $-0.0215$ with $\Delta\lambda_2=-0.0036$.
Our NFN-guided and head-gated variant further reduces the outlier fraction to $-0.0312$ and produces a substantially larger change in $\lambda_2$, indicating stronger mixing.
Implementation details of the bridge construction are provided in Appendix~\ref{secA:bridge_details} in the supplementary material.
\begin{table}[t]
\centering
\caption{\textbf{HCT decoding with different global priors.}
All methods follow the same decoding protocol (VQ-LL-32 codebook, 1000-seed CLIP top-1 association, token optimization) and differ only in the injected prior at test time.
Lower is better for FID-5k, while higher is better for IS, CLIP, and SigLIP.
\textbf{Token opt.} updates only the injected global token $z_0$ while keeping $\mathbf{z}_{1:T}$ fixed. The standard HCT optimization that updates all tokens is summarized in Appendix~\ref{subsecA:whichtoken} in the supplementary material.}

\label{tab: hct}
\resizebox{\textwidth}{!}{
\begin{tabular}{lcccc}
\toprule

\textbf{Method}    & FID-5k ($\downarrow$) & IS ($\uparrow$) & CLIP ($\uparrow$) &  SigLIP ($\uparrow$)\\

\midrule
\multicolumn{5}{l}{(DINOv2-L/14) (VQ-LL-32, 1000, CLIP top-1\%, token opt.)} \\
\midrule
HCT w/o prior ~\cite{abs-2506-08257}  & 21.2  & 281 & \cellcolor{tableorange}0.40 &3.5 \\
HCT w/ Random prior & 22.5  &  278 & \cellcolor{yellow}0.39 & 3.4 \\
HCT w/ [$\texttt{CLS}$] prior & \cellcolor{yellow}20.5  & 281 & \cellcolor{yellow}0.39 & \cellcolor{yellow}3.6 \\
HCT w/ test-time register prior~\cite{abs-2506-08010} & 21.5& \cellcolor{yellow}283 & \cellcolor{tableorange}0.40  & \cellcolor{yellow}3.6  \\
HCT w/ trained register prior~\cite{DarcetOMB24} & \cellcolor{tableorange}20.4  &  \cellcolor{tableorange}285 & \cellcolor{tablered}\textbf{0.41} & \cellcolor{tableorange}3.8 \\

HCT w/ RegToken &\cellcolor{tablered}\textbf{20.1} & \cellcolor{tablered}\textbf{289} & \cellcolor{tablered}\textbf{0.45} & \cellcolor{tablered}\textbf{3.9}\\

\midrule
\multicolumn{5}{l}{(OpenCLIP-L/14) (VQ-LL-32, 1000, CLIP top-1\%, token opt.)} \\
\midrule
HCT w/ \tok{CLS} prior & \cellcolor{yellow}20.8 & \cellcolor{yellow}283 & \cellcolor{yellow}0.40 & \cellcolor{yellow}3.5 \\
HCT w/ test-time register prior~\cite{abs-2506-08010} & \cellcolor{tableorange}21.1 & \cellcolor{tableorange}284 & \cellcolor{tableorange}0.41 & \cellcolor{tableorange}3.7 \\
HCT w/ RegToken & \cellcolor{tablered}\textbf{20.3} & \cellcolor{tablered}\textbf{287} & \cellcolor{tablered}\textbf{0.42} & \cellcolor{tablered}\textbf{3.8} \\

\bottomrule
\end{tabular}
} %
\end{table}
Across 1000 random seeds, RegToken yields consistently better optimization trajectories with lower variance (Table~\ref{tab:opt_dynamics_summary} in the supplementary material), suggesting the improvements are not driven by a small subset of favorable initializations.

\begin{figure*}[t!]
    \centering
    \includegraphics[width=\textwidth]{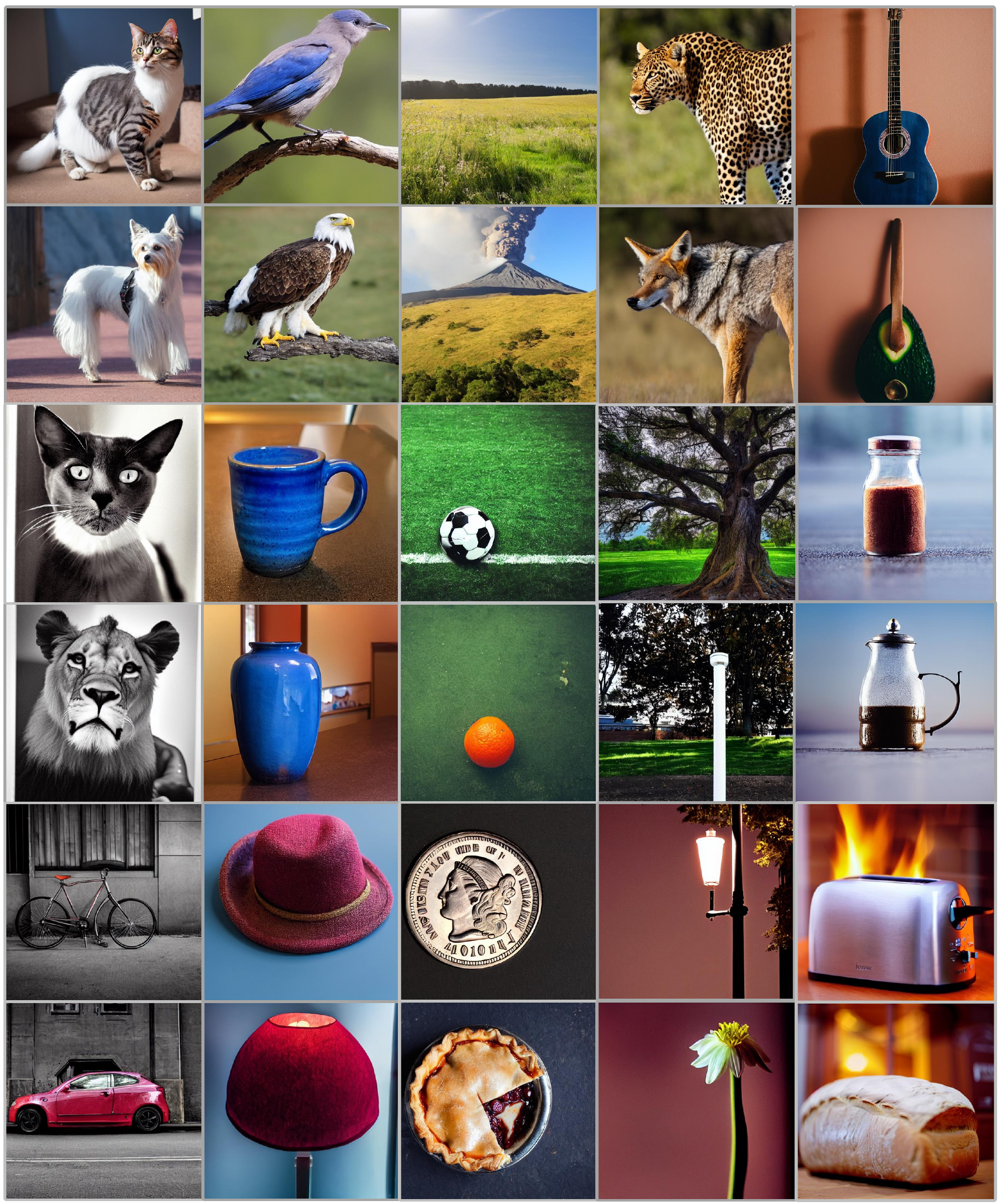}

\caption{\textbf{Qualitative results.}
Following the HCT protocol, register-based priors generate diverse images from one input image using the prompt ``\texttt{a photo of the [class]}''.
Every two rows form a group: the upper row shows input images and the lower row shows the corresponding generations.}
    \label{fig: gen_2026}
\end{figure*}

\subsection{Plug-and-play Registers for Generation}

\myparagraph{Setting and Metrics.}
We use a ViT-L/14 backbone (DINOv2) and extract register features from three NFN-selected layers near the onset of outliers (typically $\{\ell,\ell-1,\ell-2\}$).
Register vectors are projected into the HCT/TiTok-style code space through lightweight alignment, either whitening followed by nearest-neighbor search or a linear Procrustes mapping. At decoding time we inject a single global prior using two strategies.
\emph{Init-prior} mixes the first code with strength $\gamma\in\{0.25,0.5,0.75,1.0\}$.
\emph{Soft-bias} reweights token probabilities using a cosine prior
\begin{equation}
p'(c) \propto p(c) \cdot \exp\big(\beta \cdot \cos(e_c,\hat{u})\big),
\end{equation}
where $\beta\in\{0.5,1,2\}$ and $e_c$ is the code embedding.
Results use validation setting $\beta=1$.
Additional equations and hyperparameters appear in Appendices~\ref{subsecA:prior} and~\ref{subsecA:reprodetail} in the supplementary material.

\begin{table}[t]
\centering
\captionof{table}{\textbf{Effect of Prior Source (same/cross/random).}
All settings use the register pipeline and HCT decoding; only the prior source differs.}

\label{tab:prior_source}
%\resizebox{0.97\textwidth}{!}{
\begin{tabular}{lcccc}
\toprule
\textbf{Prior source} & FID-5k ($\downarrow$) & IS ($\uparrow$) & CLIP ($\uparrow$) & SigLIP ($\uparrow$) \\
\midrule
Same-image register           & 20.1 & 289 & 0.45 & 3.9 \\
Cross-image (same class)      & 20.6 & 285 & 0.41 & 3.8 \\
Random-image (any class)      & 21.1 & 283 & 0.40 & 3.7 \\
\bottomrule
\end{tabular}
%} %
\end{table}

\myparagraph{Baselines.}
We compare six variants under the same backbone, tokenizer, and decoding configuration:
\emph{no prior}~\cite{abs-2506-08257},
\emph{random prior},
\texttt{[CLS]} \emph{prior},
\emph{test-time register prior}~\cite{abs-2506-08010},
\emph{trained register prior}~\cite{DarcetOMB24},
and \emph{ours} (NFN + TokenRank + interpolation).
Detailed definitions are given in Appendix~\ref{subsecA:reprodetail} in the supplementary material.

\myparagraph{Results.}
Table~\ref{tab: hct} reports quantitative results under the token-optimization setting, and Figure~\ref{fig: gen_2026} shows qualitative examples.
Relative to the no-prior HCT baseline, all meaningful priors (\texttt{[CLS]}, test-time registers, trained registers, and ours) improve FID, IS, and text–image alignment, indicating that a single global token provides a useful handle for 1D token generation.
Random priors, by contrast, slightly degrade performance.
Among the learned or test-time variants, the trained-register prior~\cite{DarcetOMB24} provides a strong baseline due to additional training on explicit register tokens.
Nevertheless, our training-free prior achieves the best overall trade-off, obtaining the lowest FID-5k and consistently higher IS, CLIP, and SigLIP scores.
Beyond final metrics, the proposed prior also improves optimization dynamics. The OpenCLIP backbone exhibits the same ranking among priors (RegToken $\ge$ TTR $\ge$ CLS), with slightly smaller absolute gains than DINOv2, suggesting the proposed prior is transferable across backbone families while benefiting from stronger backbone--codebook alignment.
It reaches the same CLIPScore threshold in fewer steps (Steps@$\tau$: $74\rightarrow52$) and yields higher optimization AUC with lower variance across seeds (Table~\ref{tab:opt_dynamics_summary} in the supplementary material).

\myparagraph{Effect of Prior Source.}
To avoid ambiguity, we treat the same-image prior as a source-conditioned diagnostic setting rather than a text-only zero-shot generation protocol. It provides an upper-bound analysis of whether register-derived global information is useful to the frozen decoder. To examine whether the prior is transferable
rather than merely privileged information from the target image, we vary its source while keeping the pipeline and HCT decoding fixed. Table~\ref{tab:prior_source} compares registers extracted from the same image, from another image of the same class, and from random images. Same-image registers perform best, while cross-image same-class registers remain competitive. Fully random-image priors noticeably degrade quality. These results suggest that the register prior carries class- and image-level statistics that benefit tokenized generation, rather than simply acting as an arbitrary extra token. Additional generality checks are provided in Appendix~\ref{secA:generality} of the supplementary material. Across compact 1D token budgets and tokenizer variants, RegToken consistently improves over the corresponding no-prior and \texttt{[CLS]}-prior baselines. In a MaskGIT-VQGAN~\cite{ChangZJLF22} pilot, using RegToken as a frozen codebook-logit initialization bias improves FID-1k from 31.8 / 31.1 for baseline / \texttt{[CLS]} initialization to 30.3. Across DINOv2-B/L/g, RegToken also remains stronger than the corresponding \texttt{[CLS]} prior. Additional ablations in Appendix~\ref{app:ablation} show stable performance across register-neuron counts, scaling factors, NFN layer choices, outlier-token counts, and soft-bias strengths.

\section{Conclusion}
\label{sec:conclu}
We show that test-time registers can serve as global priors for token-based image generation.
We introduced {RegToken}, a training-free method that extracts register-associated structure from frozen vision transformers and converts it into a small set of reusable tokens. When used with a compact 1D token generation pipeline, these tokens consistently
improve visual quality and text--image alignment while keeping both the backbone and decoder fixed. More broadly, our findings suggest that structures previously viewed as attention artifacts can instead provide compact global context for generative models.

\bibliographystyle{splncs04}
\bibliography{main}

% ============================================================================
% Supplementary material 
% ============================================================================
\clearpage
\phantomsection
\begin{center}
  {\LARGE\bfseries Supplementary Material\par}
\end{center}
\vspace{1.0em}
% APPENDIX
%%%%%%%%%%%%%%%%%%%%%%%%%%%%%%%%%%%%%%%%%%%%%%%%%%%%%%%%%%%%%%%%%%%%%%%%%%%%%%%
%%%%%%%%%%%%%%%%%%%%%%%%%%%%%%%%%%%%%%%%%%%%%%%%%%%%%%%%%%%%%%%%%%%%%%%%%%%%%%%

%\clearpage
\appendix
%\onecolumn

\providecommand*{\theHpage}{\arabic{page}}
\providecommand*{\theHsection}{\arabic{section}}
\providecommand*{\theHsubsection}{\arabic{section}.\arabic{subsection}}

\renewcommand*{\theHpage}{A\arabic{page}}
\renewcommand*{\theHsection}{A\arabic{section}}
\renewcommand*{\theHsubsection}{A\arabic{section}.\arabic{subsection}}

%\clearpage
\phantomsection

\phantomsection

%\section*{Supplementary Material Contents}

\hypersetup{linkcolor=eccvblue}
% SectionA
\appitem{Generation Protocol and Evaluation Details}{app:generation_details}
\appsubitem{Prior injection, Prior + Opt, and Standard HCT optimization}{subsecA:prior}
\appsubitem{Which Tokens Are Fixed and Which Are Optimized}{subsecA:whichtoken}
\appsubitem{Steps@$\tau$, AUC, Thresholds, and Evaluation Settings}{subsecA:steps}
\appsubitem{Reproducibility Details}{subsecA:reprodetail}
% SectionB
\appitem{Additional Qualitative Comparisons}{app:qualitative}
\appsubitem{Same-image Prior Transfer}{subsecA:sameimg}
\appsubitem{Same-class / Cross-image Prior Transfer}{subsecA:samecls}
\appsubitem{Comparison with Different Priors}{subsecA:compari}
\appsubitem{Additional Decoding Examples Across Prompts/Classes}{subsecA:addideco}
% SectionC
\appitem{Causal Validation of the Proposed Prior}{app:causal}
\appsubitem{Low-frequency vs. High-frequency Components}{secA:lvsh}
\appsubitem{Shuffled / Matched-norm / Random Controls}{secA:shuffled}
\appsubitem{Source Sensitivity Analysis}{secA:sourcesen}
% SectionD
\appitem{Additional Analysis of Register Structure}{secA:add_ana}
\appsubitem{NFN-based Layer/Module Localization}{secA:NFN_local}
\appsubitem{NFN-based Localization on OpenCLIP}{secA:NFN_layer}
\appsubitem{Bridge: Token-spectrum vs. Pixel DCT Low-frequency}{secA:bridge_dct}
\appsubitem{Optimization Dynamics Summary}{secA:opt_dyn}
% SectionE
\appitem{Additional Ablations}{app:ablation}
\appsubitem{Number of Selected Register Neurons}{secA:num_reg_neurons}
\appsubitem{Effect of NFN-based Layer Selection}{secA:ab_nfn}
\appsubitem{Effect of TokenRank Head Gating}{secA:ab_tokenrank}
\appsubitem{Effect of Interpolation / Conservation Update}{secA:ab_interp}
\appsubitem{Details for the Value-to-TokenRank Bridge}{secA:bridge_details}
\appsubitem{Hyperparameter Sensitivity}{secA:hyperparam_sensitivity}
\appsubitem{Fusion with [CLS]}{secA:cls_fusion}
% SectionF
\appitem{Component Summary and Practical Interpretation}{secA:component_summary}
\appsubitem{What Each Component Contributes}{secA:component_contrib}
\appsubitem{Which Failure Mode Each Component Addresses}{secA:component_failure_modes}
\appsubitem{Minimal Working Configuration}{secA:minimal_config}
% SectionG
\appitem{Linear Probe on ImageNet}{secA:linear_image}
% SectionH
\appitem{Failure Cases and Limitations}{secA:limit}
\appsubitem{Failure Examples}{secA:failure_examples}
\appsubitem{Cases Where the Prior Helps Less}{secA:helps_less}
\appsubitem{Scope and Generalization Limits}{secA:scope_limits}
% SectionI
\appitem{Algorithms and Additional Generality Checks}{secA:algor}
\appsubitem{Generality Beyond the Main HCT-style Setting}{secA:generality}
\hypersetup{linkcolor=pearThree}
\clearpage

% Keep continuous PDF page numbering in the combined arXiv version.

\renewcommand{\thefigure}{S\arabic{figure}}
\setcounter{figure}{0}
\renewcommand{\thetable}{S\arabic{table}}
\setcounter{table}{0}

\section{Generation Protocol and Evaluation Details}
\label{app:generation_details}

\subsection{Prior Injection, Prior + Opt, and Standard HCT Optimization}
\label{subsecA:prior}
Backbone and decoder weights remain frozen throughout. Here, ``inserted prior''
denotes the register prior token(s) injected into the decoder token sequence.
We consider three decoding protocols: prior injection, Prior + opt, and standard
HCT optimization. In prior injection, the register prior token(s) are inserted
into the decoder token sequence without any token updates. In Prior + opt, only
the inserted global token is optimized while keeping the remaining tokens fixed.
Standard HCT optimization instead updates all token variables. For the soft-bias variant, we add a cosine prior at the logits,
\begin{equation}
l_c' = l_c + \beta \cdot \cos(e_c, \hat{u}),\qquad
\hat{u} = \frac{W z_{\mathrm{reg}}}{\lVert W z_{\mathrm{reg}}\rVert},
\end{equation}
equivalently,
\begin{equation}
p'(c) \propto p(c)\cdot \exp\big(\beta\cdot\cos(e_c,\hat{u})\big).
\end{equation}
When token optimization is enabled, we update only the designated global slot
$z_0$ while keeping $\mathbf{z}_{1:T}$ fixed:
\begin{equation}
\min_{z_0}\ \mathcal{L}_\mathrm{CLIP}\!\big(\mathcal{D}([z_0,\mathbf{z}_{1:T}]),\,\text{text}\big)
+\lambda\|z_0-\tilde r\|_2^2 .
\end{equation}

\subsection{Which Tokens Are Fixed and Which Are Optimized}
\label{subsecA:whichtoken}
\Cref{tab:gen_protocol} summarizes the optimization variables for each
setting. In particular, under Prior + opt, only the designated global slot
$z_0$ is updated, while $\mathbf{z}_{1:T}$ and all network weights remain frozen.

\begin{table}[t]
\centering
\small
\setlength{\tabcolsep}{4.0pt}
\renewcommand{\arraystretch}{1.05}
\caption{\textbf{Generation protocol and optimization variables.}
Comparison of generation settings indicating which tokens are fixed and which are
optimized under different decoding protocols.}
\label{tab:gen_protocol}
\begin{tabular}{lcc}
\toprule
Setting & Fixed tokens & Optimized tokens \\
\midrule
No prior & all $\mathbf{z}_{0:T}$ & none \\
Prior injection (default) & all $\mathbf{z}_{0:T}$ & none \\
Prior + opt (ours) & $\mathbf{z}_{1:T}$ & inserted token(s) only \\
Standard HCT opt (reference) & none & all $\mathbf{z}_{0:T}$ \\
\bottomrule
\end{tabular}
\end{table}

\subsection{Steps@$\tau$, AUC, Thresholds, and Evaluation Settings}
\label{subsecA:steps}
To quantify optimization efficiency and stability beyond final metrics, we summarize
CLIPScore trajectories under the Prior + opt protocol. Unless otherwise noted, we use
$S=50$ optimization steps and set $\tau$ to $0.95\times$ the final CLIPScore of the
No prior baseline at step $S$. Steps@$\tau$ measures the number of steps required to
reach the target threshold, AUC measures the area under the CLIPScore trajectory, and
Std@S reports the standard deviation across seeds at the final step. \Cref{tab:opt_dynamics_summary}
reports these summary statistics for all compared priors under the same evaluation setup.

\begin{table}[t]
\centering
\caption{\textbf{Test-time optimization efficiency and stability.}
Summary statistics of CLIPScore trajectories under the \emph{Prior + opt} protocol,
reporting the steps to reach the target threshold (Steps@$\tau$), optimization AUC,
and variance across seeds.}
\label{tab:opt_dynamics_summary}
\small
\setlength{\tabcolsep}{6pt}
\renewcommand{\arraystretch}{1.05}
\begin{tabular}{lccc}
\toprule
\textbf{Prior} & Steps@$\tau$ ($\downarrow$) & AUC ($\uparrow$) & Std@S ($\downarrow$) \\
\midrule
No prior & 74 & 0.342 & 0.018 \\
Random prior & 92 & 0.319 & 0.023 \\
\texttt{[CLS]} prior & 66 & 0.351 & 0.017 \\
Test-time register prior~\cite{abs-2506-08010} & 63 & 0.354 & 0.016 \\
Trained register prior~\cite{DarcetOMB24} & 58 & 0.360 & 0.015 \\
RegToken & \textbf{52} & \textbf{0.366} & \textbf{0.013} \\
\bottomrule
\end{tabular}
\end{table}

\subsection{Reproducibility Details}
\label{subsecA:reprodetail}
All methods share the same HCT decoding setup and differ only in the injected prior:
no prior, random prior (matched norm), \texttt{[CLS]} prior, test-time register prior,
trained register prior, and ours (NFN + TokenRank + interpolation). Unless otherwise
noted, the backbone and decoder remain frozen throughout. For the soft-bias variant,
we use $\beta\in\{0.5,1,2\}$. For Prior + opt, only the inserted global slot is updated,
using the same optimization budget of $S=50$ steps across all methods, with
$\lambda\in[0.1,1]$ and step size $\eta\in[1\times10^{-3},\,5\times10^{-3}]$.
Generation metrics are evaluated on the same ImageNet-\texttt{val} subset, while diagnostic
analyses use the same 1k-image subset unless otherwise stated. We report FID-5k, IS,
CLIP, SigLIP, and trajectory-based metrics (Steps@$\tau$, AUC, and Std@S) under
identical settings for all compared priors.

\myparagraph{Why Cross-model Transfer is Plausible.}
Although the source backbone (e.g., DINOv2 or OpenCLIP) and the target tokenizer/decoder (e.g., HCT/TiTok) are different models, the proposed prior is not intended to transfer task-specific classifier logits or exact token semantics. Instead, RegToken extracts a compact global signal dominated by smooth scene-level statistics, such as coarse layout, illumination, and global appearance, and maps it into the decoder token space through the same alignment procedure used for all priors. This makes the transfer problem closer to injecting a global structural bias than to matching model-specific semantic readouts. Empirically, the same-image and same-class transfer results in Table~\ref{tab:prior_source} and Figure~\ref{fig:source_sensitivity} support this interpretation: priors from related sources remain useful, whereas random-source priors degrade performance.

\myparagraph{Source-conditioned vs. reference-transfer settings.}
We distinguish the same-image diagnostic setting from target-free reference-transfer
settings. The same-image prior is not a pure text-only or true zero-shot generation
protocol; instead, it should be interpreted as a source-conditioned upper-bound
analysis of whether register-derived global information is useful to the frozen
decoder. In the reference-transfer setting, the prior is extracted from an independent
image. Same-class references remain useful, whereas random-source references degrade
quality, indicating transferable global structure such as layout, illumination, and
appearance rather than simple source memorization. For image-conditioned variants, input-image conditioning is part of the task
definition rather than information leakage; target images are used only for
evaluation or visualization and are never provided when constructing the prior.

\section{Additional Qualitative Comparisons}
\label{app:qualitative}

\subsection{Same-image Prior Transfer}
\label{subsecA:sameimg}
\begin{figure}[t!]
\centering
\includegraphics[width=\textwidth]{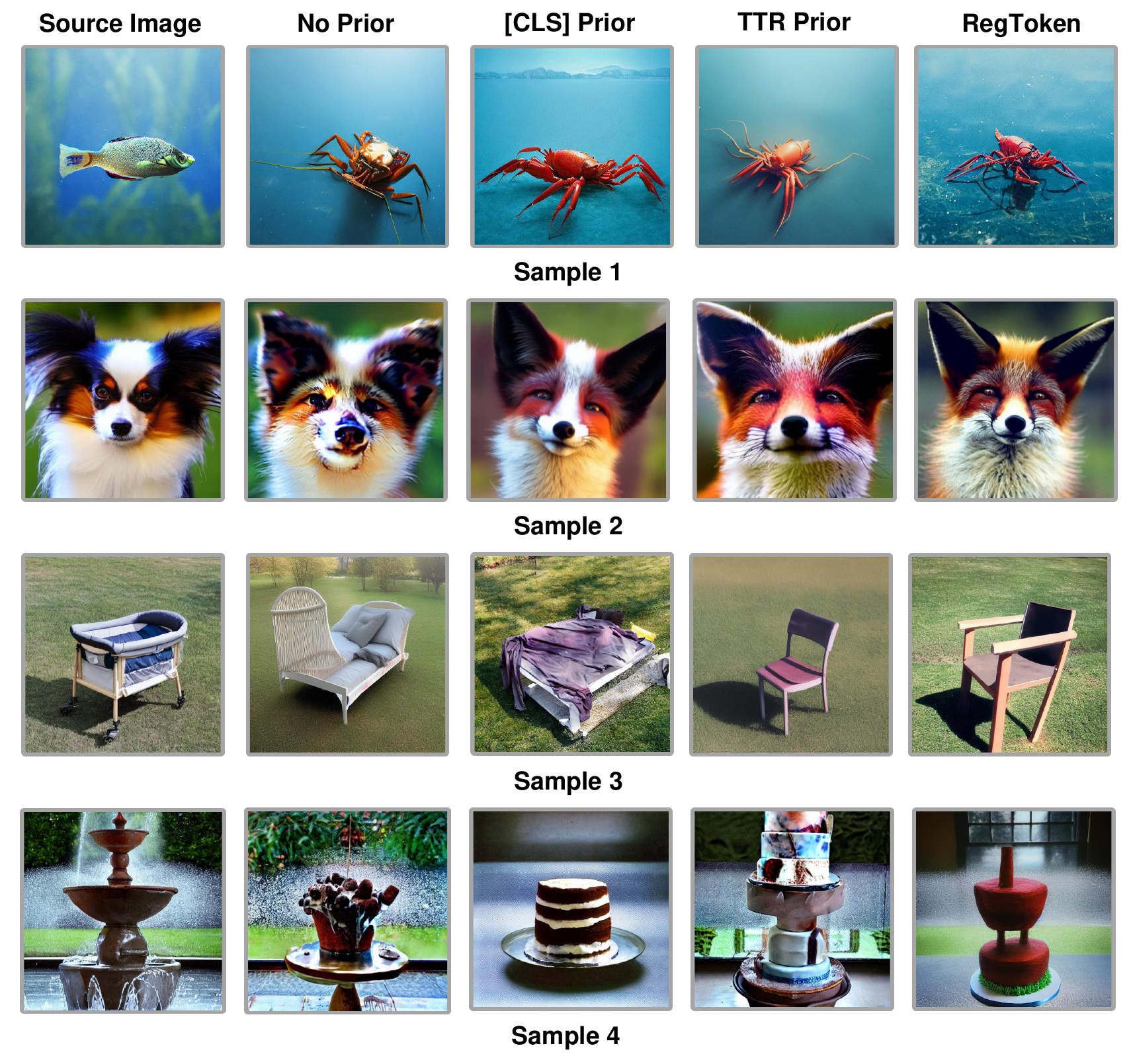}
\caption{\textbf{Same-image Prior Transfer.}
Qualitative comparison when the inserted prior is extracted from the same source image. Each row shows one source image and the decoded result for the same target prompt under different prior choices. Compared with no prior and alternative global summaries, RegToken produces a more coherent global layout and a more stable target-category rendering, while still preserving useful image-specific global characteristics from the source.}
\label{fig:qual_same_image}
\end{figure}

We first visualize the case where the inserted prior is extracted from the same source image. This setting isolates whether the prior can provide useful global structure to the frozen decoder without introducing cross-image mismatch. Compared with no prior and alternative global summaries, RegToken produces more
coherent global layout and more stable target-category rendering across examples.
\Cref{fig:qual_same_image} shows representative cases.

\subsection{Same-class / Cross-image Prior Transfer}
\label{subsecA:samecls}

\begin{figure}[t!]
\centering
\includegraphics[width=\textwidth]{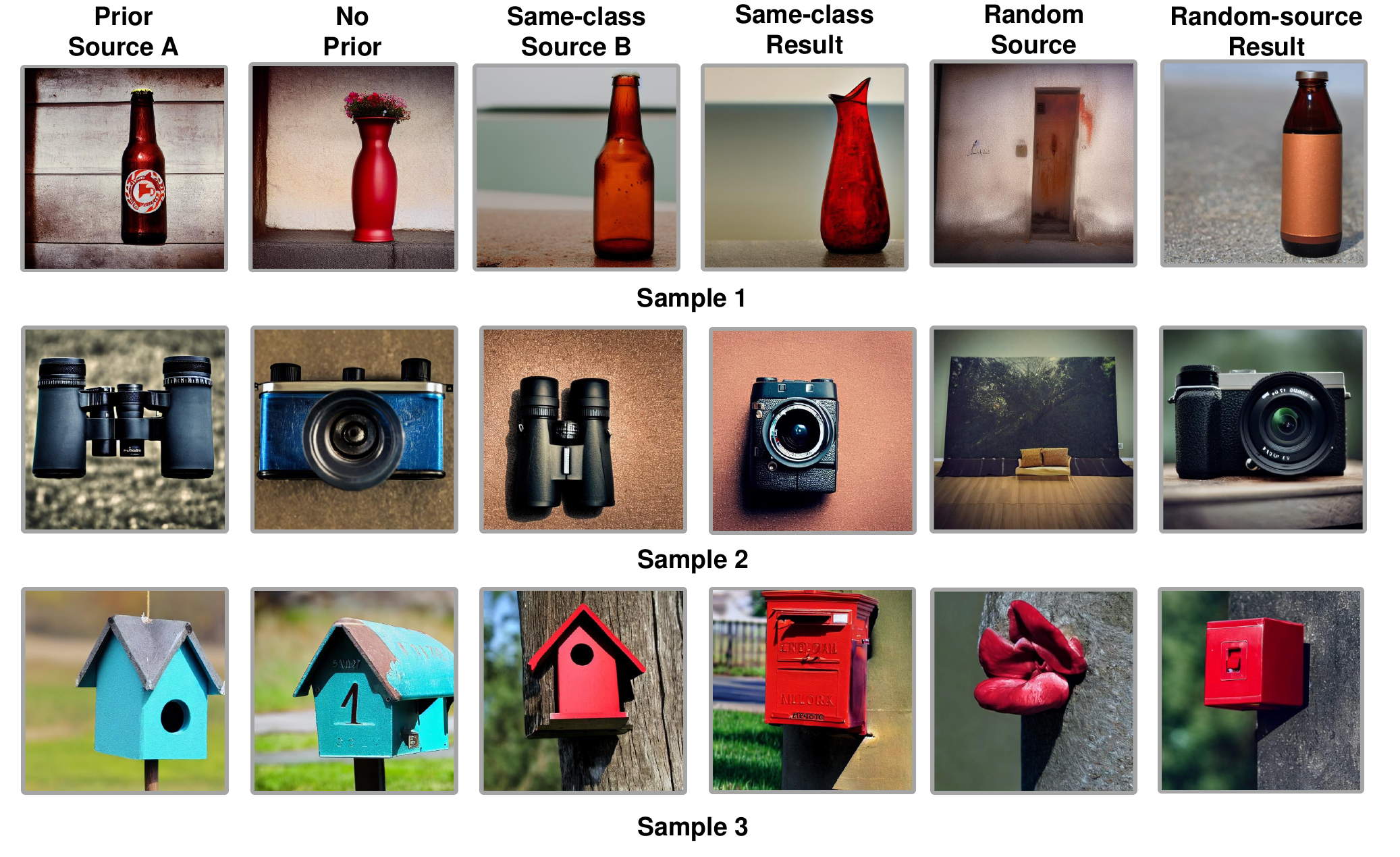}
\caption{\textbf{Same-class and Cross-image Prior Transfer.}
Each row compares different prior-source relations under the same target prompt.
From left to right, we show the original prior source image, the no-prior result,
a different source image from the same semantic class, the corresponding same-class
transfer result, a random cross-class source image, and the corresponding random-source
transfer result. Same-class transfer generally preserves more useful source-derived
global structure than random-source transfer, while still producing the target category,
indicating that the proposed prior captures genuinely transferable global information
rather than merely memorizing a specific source image.}
\label{fig:qual_cross_image}
\end{figure}

We next test whether the proposed prior remains useful when transferred across
images. We consider priors extracted from different source images within the
same semantic class, as well as random cross-class source images, while keeping
the target prompt fixed within each row. These examples help distinguish genuinely
transferable global structure from trivial memorization of a specific source image.
\Cref{fig:qual_cross_image} summarizes representative cases.

\subsection{Comparison with Different Priors}
\label{subsecA:compari}

\begin{figure}[t!]
\centering
\includegraphics[width=\textwidth]{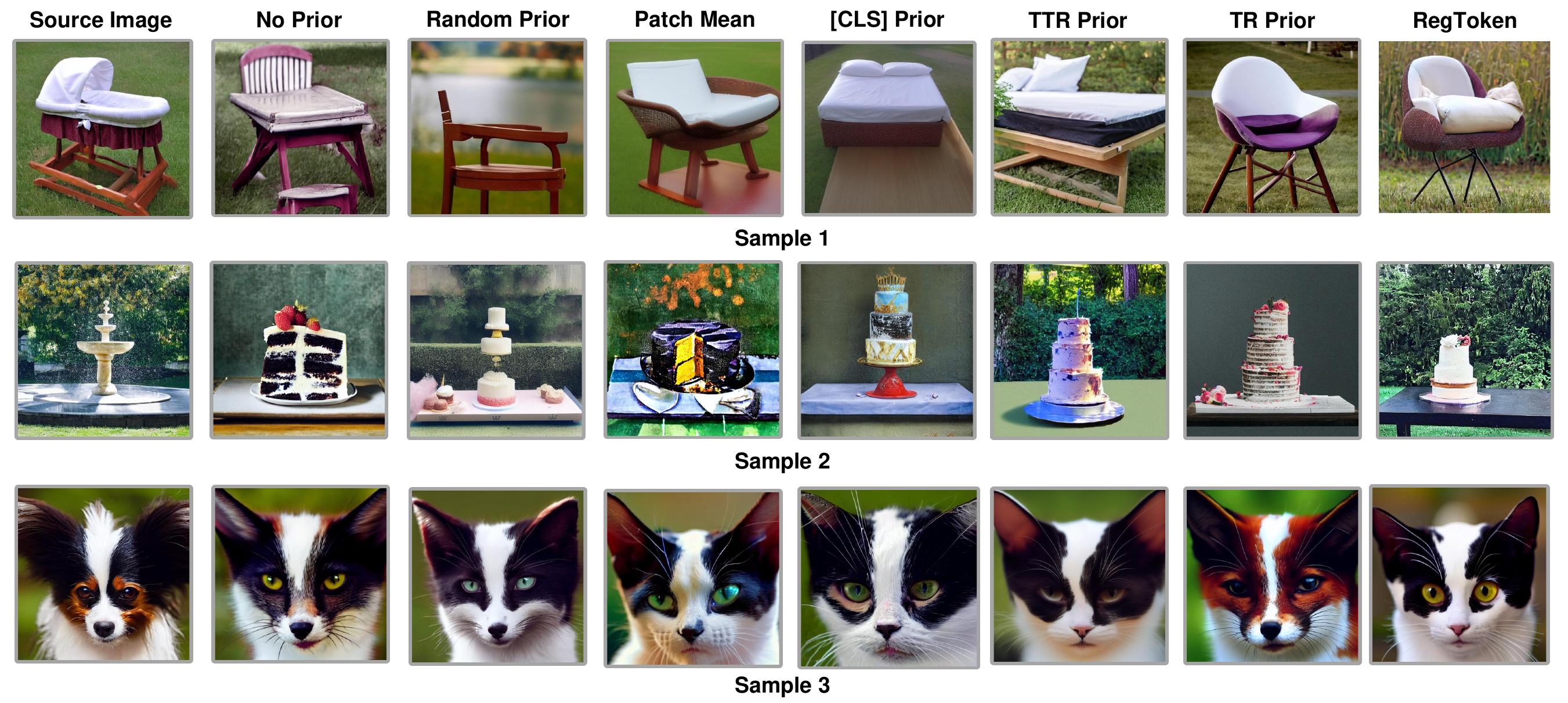}
\caption{\textbf{Comparison with Different Priors.}
Qualitative comparison under the same-image prior setting using different global
summaries and prior choices. From left to right, we show the source image, no-prior
decoding, matched-norm random prior, patch-mean prior, \texttt{[CLS]} prior,
test-time register prior, trained register prior, and RegToken. Compared with
simpler global summaries and alternative prior constructions, RegToken produces
more coherent target-category rendering while better preserving useful source-derived
global structure, indicating that the gain comes from the specific prior extracted
by the proposed NFN + TokenRank pipeline rather than from generic token injection.}
\label{fig:qual_baselines}
\end{figure}

We further compare RegToken against alternative global summaries and prior choices
under the same-image prior setting. Specifically, we compare matched-norm random
priors, patch-mean priors, \texttt{[CLS]} priors, test-time registers, and trained
registers. These comparisons clarify that the gain does not come from merely
injecting an additional token, but from the specific structure extracted by the
proposed NFN + TokenRank pipeline. \Cref{fig:qual_baselines} provides side-by-side
examples.

\subsection{Additional Decoding Examples Across Prompts/Classes}
\label{subsecA:addideco}

\begin{figure}[t!]
\centering
\includegraphics[width=\textwidth]{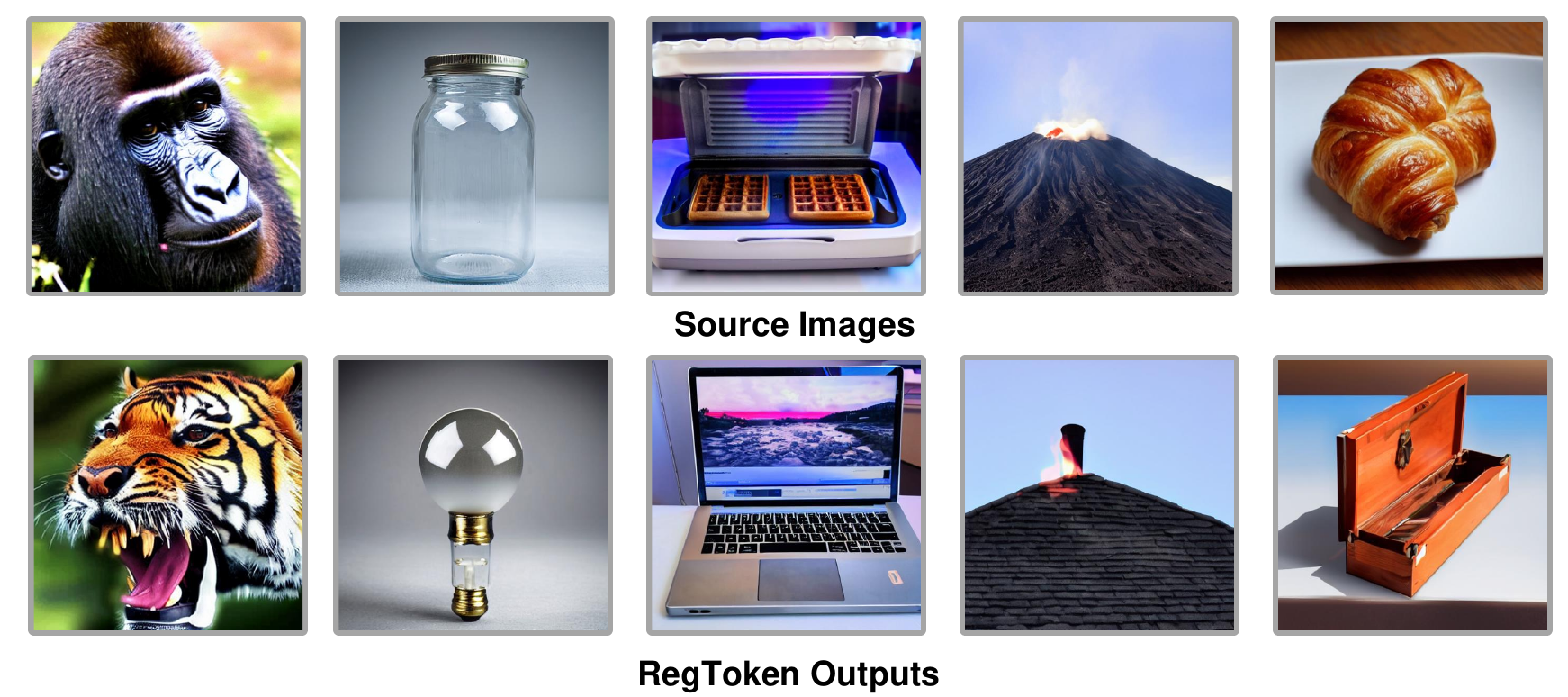}
\caption{\textbf{Additional Decoding Examples Across Prompts and Classes.}
A small gallery of additional examples beyond the representative cases shown in
the main paper and earlier appendix figures. The top row shows source images,
and the bottom row shows the corresponding RegToken outputs under different target
prompts. These examples illustrate the breadth of the proposed prior across diverse
semantic classes under the same frozen-decoder setting.}
\label{fig:qual_additional}
\end{figure}

Finally, we provide a small gallery of additional decoding examples across diverse
prompts and semantic classes beyond the representative cases shown in the main paper
and preceding appendix figures. Since~\Cref{fig:qual_same_image,fig:qual_cross_image,fig:qual_baselines}
already compare prior-source relations and prior families in detail, this subsection
focuses on qualitative breadth.~\Cref{fig:qual_additional} shows additional source--target
examples illustrating that RegToken generally maintains coherent global structure and
target-category consistency under the same frozen-decoder setting.

\section{Causal Validation of the Proposed Prior}
\label{app:causal}

To complement the descriptive and correlational analyses in the main paper and
\Cref{secA:add_ana}, we provide additional controlled comparisons to test whether
the gain of RegToken comes from the specific structure of the extracted prior,
rather than from simply injecting an extra token or adding arbitrary global bias.
We focus on three aspects: frequency decomposition, shuffled/matched controls,
and source sensitivity under prior transfer.

\subsection{Low-frequency vs. High-frequency Components}
\label{secA:lvsh}

\begin{figure}[t]
\centering
\includegraphics[width=0.85\textwidth]{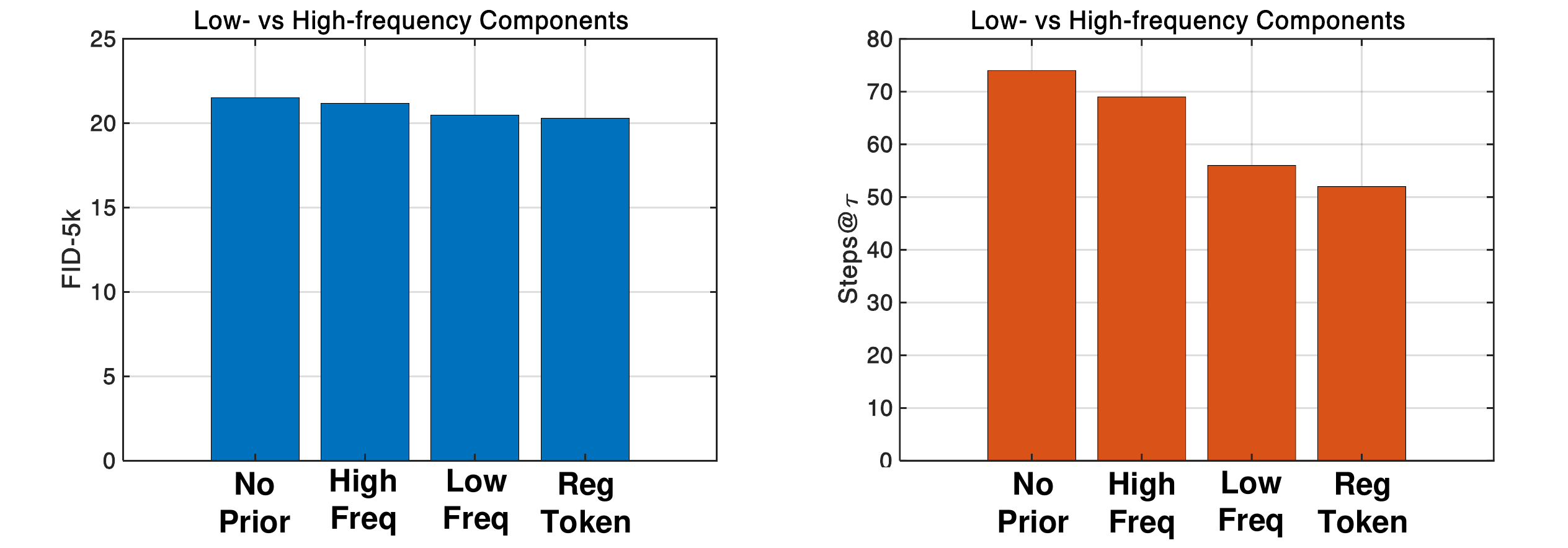}
%\vspace{-2mm}
\caption{\textbf{Low- vs. High-Frequency Components of the Proposed Prior.}
We decompose the inserted prior in the embedding domain into low- and high-frequency
components using a 1D DCT over the feature dimension, and evaluate each component
under the same decoding protocol. Left: generation quality measured by FID-5k.
Right: optimization efficiency measured by Steps@$\tau$. The low-frequency component
recovers most of the gain of the full RegToken prior, whereas the high-frequency
component yields only limited improvement over the no-prior baseline. These results
suggest that a substantial part of the benefit comes from the smoother global structure
of the proposed prior, while the full prior remains the strongest overall. We set
$k=16$ in all experiments.}
\label{fig:lvsh}
\end{figure}

To test whether the gain is associated with specific frequency components of the
proposed prior, we decompose the inserted prior in the embedding domain using a
1D DCT over the feature dimension. The first $k$ coefficients are retained as the
low-frequency component, while the remaining coefficients form the high-frequency
component. Each component is transformed back to the original embedding space and
evaluated under the same decoding setup. This comparison isolates whether the
observed improvement is mainly driven by the smoother global structure of the prior
or by high-frequency residual information. We report both generation-quality and
optimization-efficiency metrics under the same protocol described in~\Cref{app:generation_details}. Consistent trends are observed in both generation
quality and optimization efficiency, indicating that the smoother component alone
already explains a substantial portion of the improvement, although the complete
prior still performs best. The quantitative comparison is summarized in
\Cref{fig:lvsh}.

\subsection{Shuffled / Matched-norm / Random Controls}
\label{secA:shuffled}

\begin{figure}[t]
\centering
\includegraphics[width=0.75\textwidth]{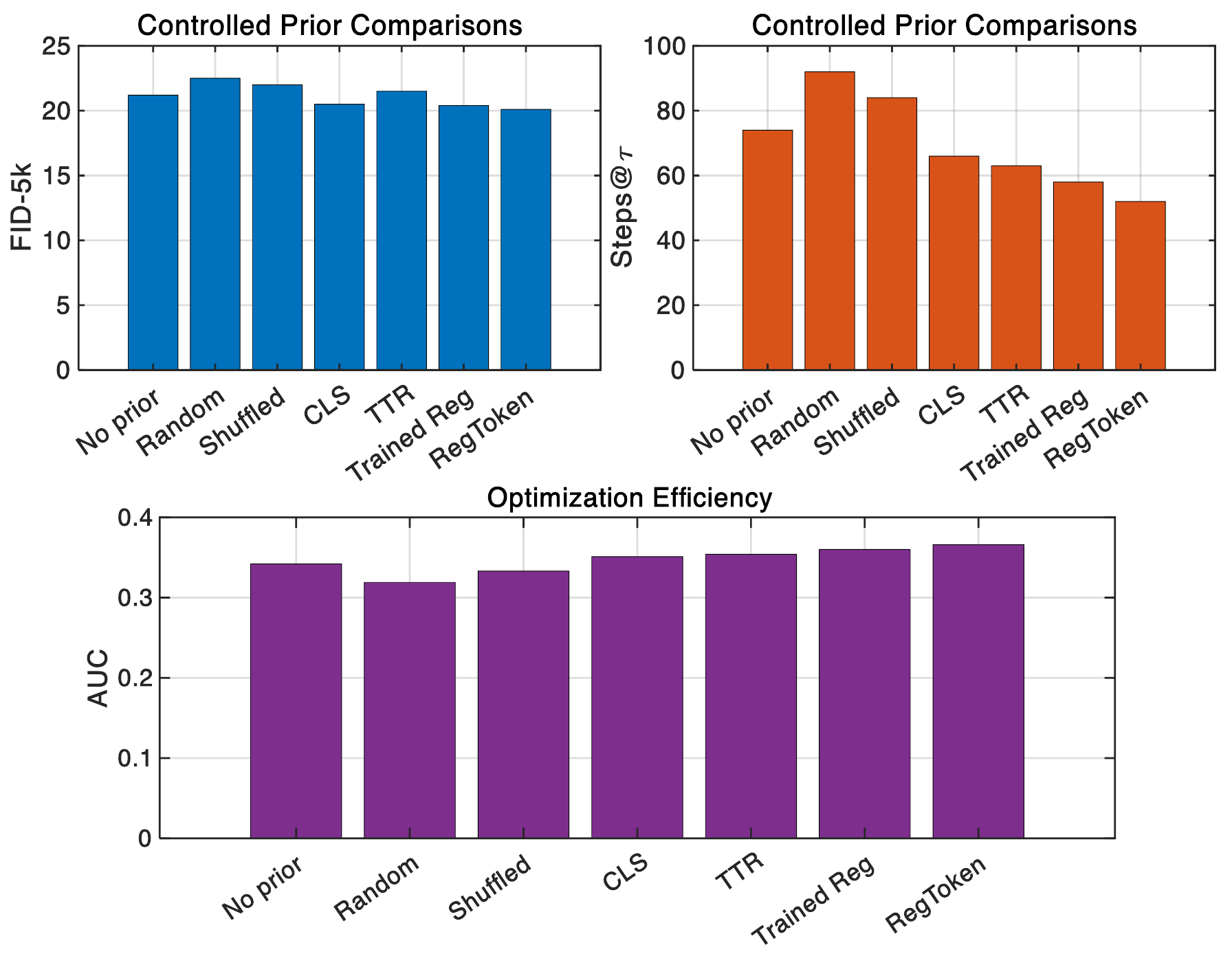}
%\vspace{-2mm}
\caption{\textbf{Controlled Comparisons with Shuffled and Alternative Priors.}
We compare RegToken with several controlled alternatives under the same decoding
protocol, including a matched-norm random prior, a shuffled prior obtained by
permuting the feature dimensions of the extracted prior, a \texttt{[CLS]} prior,
a test-time register prior, and a trained register prior. Left: generation quality
measured by FID-5k. Right: optimization efficiency measured by Steps@$\tau$.
RegToken consistently outperforms the shuffled and random controls, indicating that
the gain does not come from simply injecting an additional token of similar magnitude,
but from the structured content of the proposed prior.}
\label{fig:shuffled_controls}
\end{figure}

We next compare the proposed prior with several controlled alternatives to test
whether the gain is merely due to injecting an additional token with similar
magnitude. In particular, we consider a matched-norm random prior and shuffled
versions of the extracted prior, and compare them against the full RegToken prior
under the same decoding protocol. These controls help distinguish structured prior
content from trivial token injection effects. Here, the shuffled prior is constructed
by randomly permuting the feature dimensions of the extracted prior while preserving
its $\ell_2$ norm. We report both generation-quality and optimization-efficiency
metrics under the same protocol described in~\Cref{app:generation_details}. The
controlled comparison is summarized in~\Cref{fig:shuffled_controls}.

\subsection{Source Sensitivity Analysis}
\label{secA:sourcesen}

\begin{figure}[t]
\centering
\includegraphics[width=0.8\textwidth]{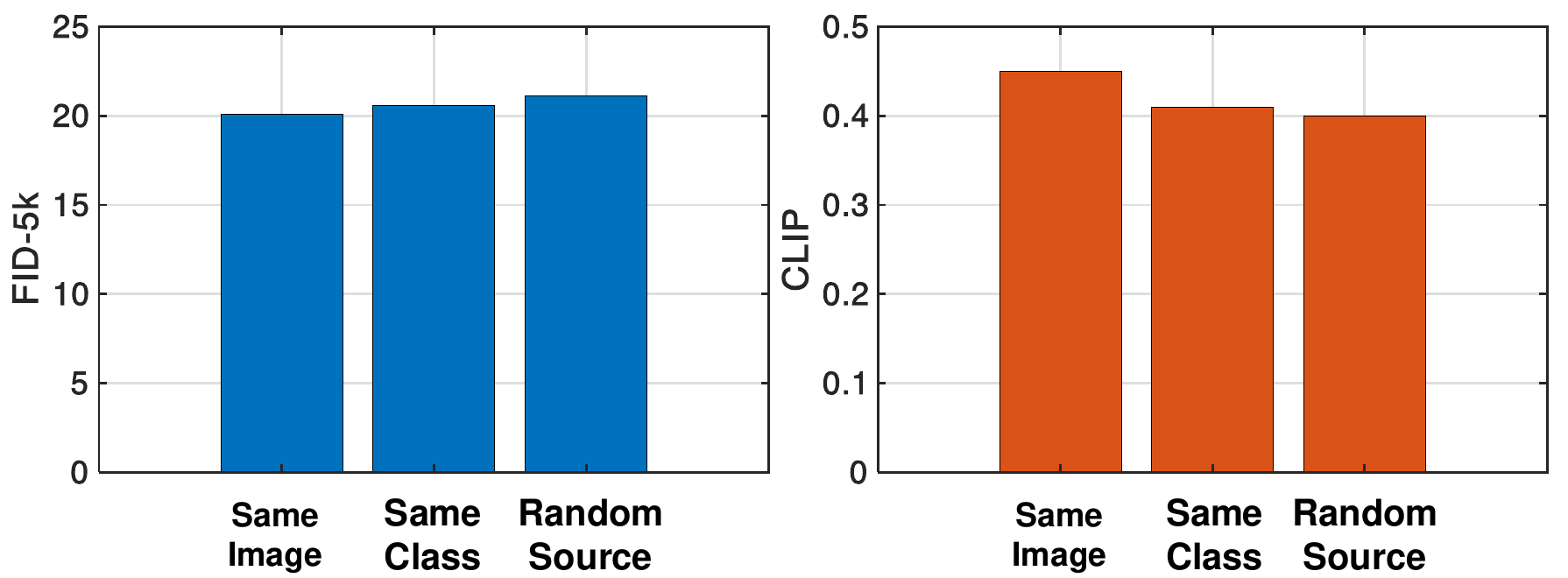}
\caption{\textbf{Source Sensitivity of the Proposed Prior.}
We compare three source conditions under the same decoding protocol: a prior extracted
from the same image, a prior transferred from a different image of the same class, and
a prior transferred from a random image. Left: generation quality measured by FID-5k.
Right: text-image alignment measured by CLIP. Same-image transfer performs best, while
same-class cross-image transfer still retains a clear advantage over random-source
transfer, suggesting that the proposed prior contains partially transferable global
structure beyond source-specific alignment.}
\label{fig:source_sensitivity}
\end{figure}

We study how sensitive the proposed prior is to the choice of source image.
Specifically, we compare same-image transfer, same-class cross-image transfer,
and random-source transfer under the same decoding setup. These results help
characterize how much of the benefit comes from transferable global structure
versus source-specific alignment. The quantitative results are summarized in
\Cref{fig:source_sensitivity} and correspond to the same source conditions
reported in the main paper (Table~\ref{tab:prior_source}).

\section{Additional Analysis of Register Structure}
\label{secA:add_ana}

\subsection{NFN-based Layer/Module Localization}
\label{secA:NFN_local}

\begin{figure*}[ht]
\centering
\includegraphics[width=0.99\textwidth]{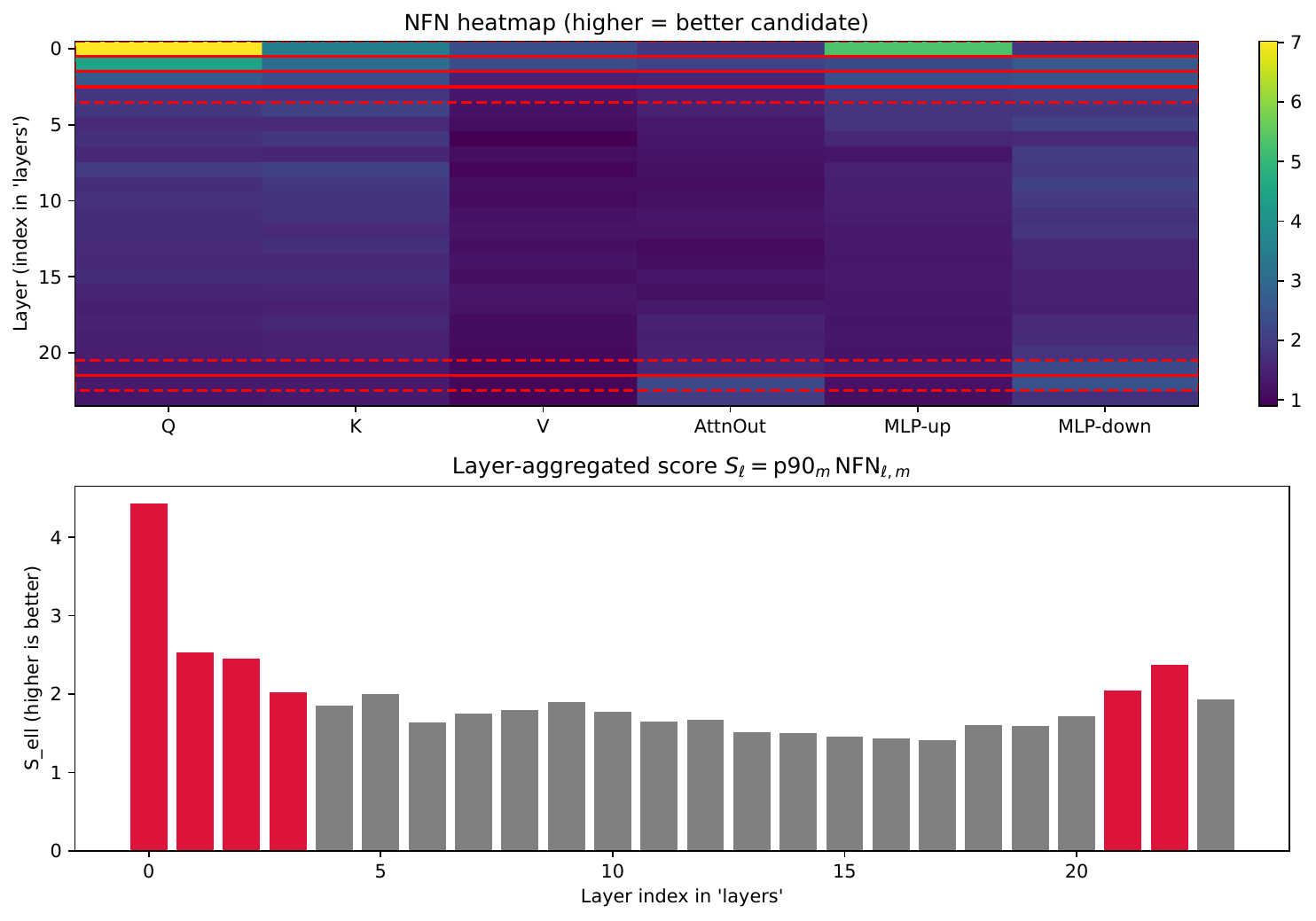}
%\vspace{-3mm}
\caption{\textbf{NFN-based layer localization in DINOv2.}
The heatmap shows NFN scores across layers and modules, computed on a 1k-image subset of ImageNet-\texttt{val}.
Layers with the largest scores are selected as candidate intervention points, which consistently align with the emergence of sink/register behavior.}
\label{fig:nfn_position}
\end{figure*}

We compute NFN scores for each module and then aggregate them by
$S_\ell=\max_m \mathrm{NFN}_{\ell,m}$ over a 1k-image random subset of
ImageNet-\texttt{val}, following the procedure described in
Section~\ref{sec:token2neurons}. The $\kappa=5$ layers with the
largest $S_\ell$ are selected as candidate insertion locations.
\Cref{fig:nfn_position} visualizes the resulting layer$\times$module heatmap
for DINOv2 and the identified candidate layers.

\subsection{NFN-based Layer Localization on OpenCLIP}
\label{secA:NFN_layer}

\begin{figure*}[ht]
\centering
\includegraphics[width=0.99\textwidth]{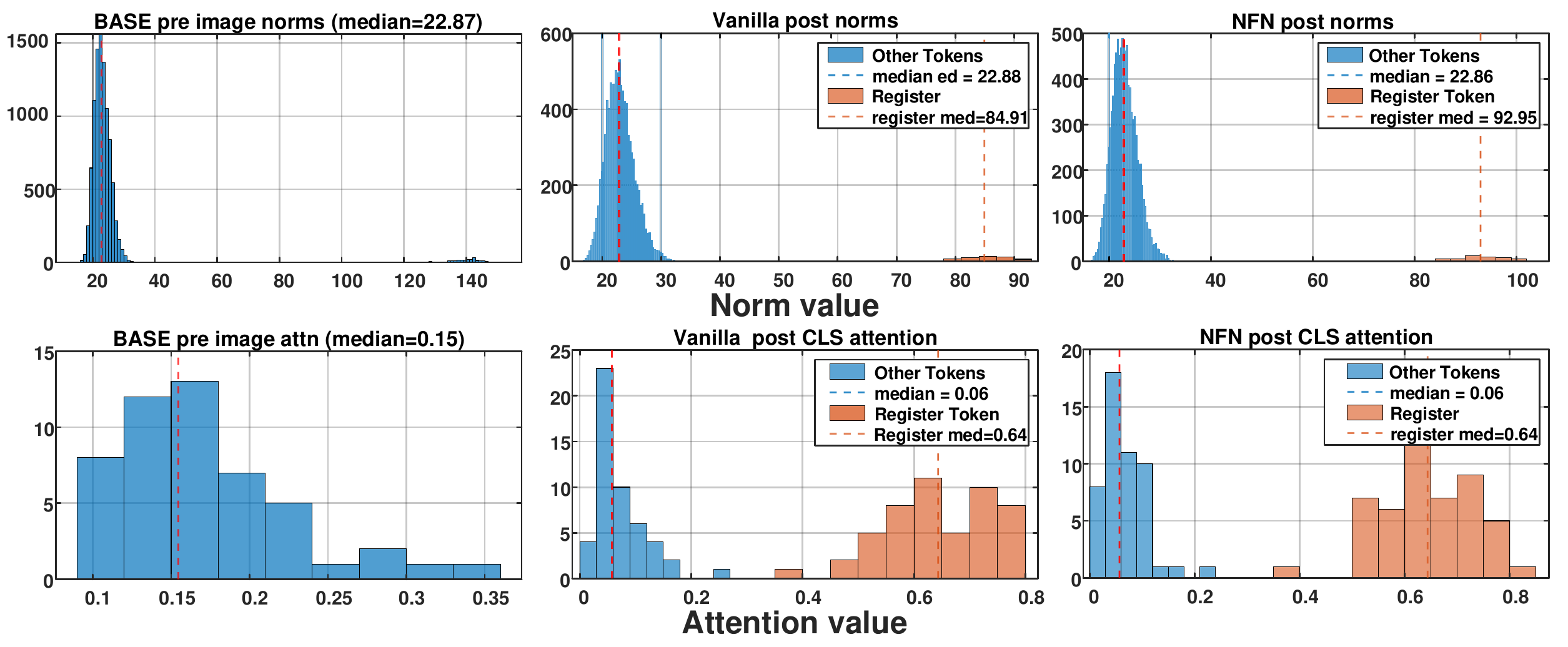}
%\vspace{-2mm}
\caption{\textbf{Effect of register insertion on OpenCLIP features.} We insert a single register token at NFN-selected layers using the top-50 register neurons.  The register token accumulates larger feature norms and attention mass, while the distributions of image tokens remain largely unchanged.}
\label{fig:nfn_clip}
\end{figure*}

OpenCLIP exhibits behavior similar to DINOv2. As shown in \Cref{fig:nfn_clip},
the median register value norm increases from 84.91 to 92.95 (+10.5\%) on the
same 1k-image ImageNet-\texttt{val} subset. In contrast, the median
CLS$\rightarrow$register attention and the corresponding statistics for image
tokens remain largely unchanged.

\subsection{Bridge: Token-spectrum vs. Pixel DCT Low-frequency}
\label{secA:bridge_dct}

To relate the embedding-spectrum diagnostic to pixel-space statistics, we
correlate it with an image-level low-frequency energy ratio
(\Cref{tab:bridge_dct_corr}). Specifically, we compute
$\mathrm{LF}_{\mathrm{DCT}}$ as the energy ratio of a fixed $k\times k$
low-frequency block ($k=8$) in the 2D DCT of RGB pixels and report Spearman
correlations with two-sided $p$-values over 1000 ImageNet images.

\begin{table}[t]
\centering
\caption{\textbf{Bridge to Pixel-space Frequency.}
Spearman correlation between the image-level low-frequency energy ratio ($\mathrm{LF}_{\mathrm{DCT}}$) computed from pixel-space DCT and token-spectrum metrics derived from register features, evaluated on 1k ImageNet images.}
%\vspace{-3mm}
\label{tab:bridge_dct_corr}
\small
\setlength{\tabcolsep}{6pt}
\renewcommand{\arraystretch}{1.05}
\begin{tabular}{lcc}
\toprule
\textbf{Backbone / Metric} & Spearman $\rho$ ($\uparrow$) & $p$-value \\
\midrule
DINOv2, $\rho_{1:15}$            & 0.22 & $<10^{-10}$ \\
DINOv2, $\bar dB_{\text{low}}$   & 0.19 & $<10^{-8}$ \\
OpenCLIP, $\rho_{1:15}$          & 0.14 & $<10^{-5}$ \\
OpenCLIP, $\bar dB_{\text{low}}$ & 0.11 & $<10^{-3}$ \\
\bottomrule
\end{tabular}
\end{table}

\subsection{Optimization Dynamics Summary}
\label{secA:opt_dyn}

We further interpret the trajectory-based metrics introduced in
\Cref{app:generation_details}. Under the \emph{Prior + opt} protocol, only the
injected token is optimized while $\mathbf{z}_{1:T}$ remains fixed. Compared
with the alternative priors, RegToken reaches the target threshold in fewer
steps, achieves the largest optimization AUC, and exhibits the smallest final
variance across seeds. These trends indicate that the proposed prior improves
both optimization efficiency and stability.

\section{Additional Ablations}
\label{app:ablation}

The full configuration (bottom row of \Cref{tab:ablation_components})
provides the best overall trade-off. Starting from the TTR baseline,
adding NFN-based layer selection and TokenRank head gating progressively improves FID, IS, and SigLIP, while the final interpolation step yields the largest gain in FID-5k and text--image alignment. Performance remains stable across the hyperparameter settings in~\Cref{tab:ablation,tab:hyperparam_sensitivity}. Varying the number of register neurons, selected channels, NFN-selected layers, outlier tokens, and scaling strengths changes FID only mildly, indicating that the method is relatively stable with respect to these choices. Removing NFN-based layer selection or TokenRank head gating leads to a small but consistent degradation, showing that
both components contribute to the final performance.

\begin{table}[t]
\centering
\caption{\textbf{Component Ablation for HCT Decoding (DINOv2-L/14 + ImageNet-\texttt{val} 5k).}
Starting from the test-time register (TTR) baseline~\cite{abs-2506-08010}, we progressively add NFN-based layer selection, TokenRank head gating, and the token--neuron interpolation rule. All settings follow the same HCT decoding protocol as in the main paper.}
%\vspace{-2mm}
\label{tab:ablation_components}
\resizebox{0.97\textwidth}{!}{
\begin{tabular}{lcccc}
\toprule
\textbf{Method} & FID-5k ($\downarrow$) & IS ($\uparrow$) & CLIP ($\uparrow$) & SigLIP ($\uparrow$) \\
\midrule
TTR baseline~\cite{abs-2506-08010}                     & 21.5 & 283 & 0.40 & 3.6 \\
TTR + NFN layer selection                              & 21.3 & 284 & 0.40 & 3.7 \\
TTR + NFN + TokenRank head gating                      & 20.8 & 286 & 0.41 & 3.8 \\
TTR + NFN + TokenRank + interpolation (RegToken, HCT) & \textbf{20.3} & \textbf{287} & \textbf{0.41} & \textbf{3.9} \\
\bottomrule
\end{tabular}
}
\end{table}

\begin{table}[t]
\centering
\caption{\textbf{Ablation of Register Hyperparameters and Components (DINOv2-L/14 + HCT).}
FID-5k on ImageNet-\texttt{val} (5k images) for different numbers of register neurons $k$, register scales $s$/$\gamma$, and settings with or without NFN-based layer selection and TokenRank head gating.}
\label{tab:ablation}
\begin{tabular}{lccc}
\toprule
\textbf{Setting} & $k$ & $s$ / $\gamma$ & FID-5k ($\downarrow$) \\
\midrule
RegToken (default)       & 32 & 2.0 & \textbf{20.3} \\
\midrule
Fewer neurons            & 16 & 2.0 & 20.7 \\
More neurons             & 64 & 2.0 & 20.7 \\
\midrule
Smaller scale            & 32 & 1.0 & 20.9 \\
Larger scale             & 32 & 4.0 & 21.0 \\
\midrule
w/o NFN layer selection  & 32 & 2.0 & 21.1 \\
w/o TokenRank gating     & 32 & 2.0 & 20.9 \\
\bottomrule
\end{tabular}
\end{table}

\subsection{Number of Selected Register Neurons}
\label{secA:num_reg_neurons}

We first study the effect of the number of selected register neurons.
As summarized in \Cref{tab:ablation}, varying the number of selected neurons
from $k=16$ to $k=64$ leads to only minor changes in FID-5k, with the default
setting $k=32$ achieving the best result. This suggests that the proposed prior
is relatively stable with respect to the exact number of selected register
dimensions within a moderate range.

\subsection{Effect of NFN-based Layer Selection}
\label{secA:ab_nfn}

We next examine the role of NFN-based layer selection. Removing NFN-based layer
selection leads to a consistent degradation in generation quality, as shown in
\Cref{tab:ablation}. A similar trend is observed in the progressive component
ablation in \Cref{tab:ablation_components}, where adding NFN-based layer selection
to the TTR baseline improves FID-5k from 21.5 to 21.3. These results indicate
that selecting intervention layers based on NFN provides a small but consistent
benefit.

\subsection{Effect of TokenRank Head Gating}
\label{secA:ab_tokenrank}

We further evaluate the effect of TokenRank-based head gating. As shown in
\Cref{tab:ablation}, removing TokenRank gating degrades FID-5k from 20.3 to 20.9.
The progressive ablation in \Cref{tab:ablation_components} shows the same trend:
adding TokenRank head gating on top of NFN-based layer selection improves FID-5k
from 21.3 to 20.8. This confirms that TokenRank helps identify heads that more
effectively carry the proposed prior.

\subsection{Effect of Interpolation / Conservation Update}
\label{secA:ab_interp}

Finally, we assess the contribution of the interpolation / conservation update.
In the progressive component ablation of \Cref{tab:ablation_components}, adding
the interpolation rule on top of NFN-based layer selection and TokenRank head
gating yields the best overall performance, improving FID-5k from 20.8 to 20.3
while also slightly improving IS and SigLIP. This suggests that preserving the
structured prior during insertion is important for achieving the full gain of
RegToken.

\subsection{Details for the Value-to-TokenRank Bridge}
\label{secA:bridge_details}

This subsection describes the implementation details of the diagnostics used in
Section~\ref{sec:token2neurons}. We couple a per-token write measure
with a global centrality score to avoid interpreting raw attention in isolation.
For a head with attention $A\in\mathbb{R}^{T\times T}$ (row-stochastic) and values
$\{v_t\}$, we define the \emph{write mass} of token $t$ as:
\[
\mathrm{WriteMass}(t)=\sum_{i=1}^{T} A_{i,t}\,\|v_t\|_2^2,
\]
which better reflects how much content is written \emph{into} $t$ than attention
alone. Treating $A$ as a Markov chain, \emph{TokenRank} is the stationary distribution
$\pi^\top=\pi^\top A$, computed per head and then median-aggregated within a layer.
We quantify head mixing by the second eigenvalue $\lambda_2$ of $A$; a smaller
$\lambda_2$ corresponds to a larger spectral gap and faster mixing.

In practice, we use post-softmax attention matrices, power iteration for $\pi$,
and report per-layer medians across heads for $\pi(\texttt{REG})$,
$\mathrm{WriteMass}(\texttt{REG})$, and $\lambda_2$. To assay absorption, we scale
the register coordinates on their subspace by $s\in\{0.5,1,2,4,8,16\}$ and track
TokenRank(\texttt{REG}), WriteMass(\texttt{REG}), and the median $\lambda_2$.
Monotone increases in the first two together with a decrease in $\lambda_2$
indicate that the added token behaves as a controlled absorber rather than merely
amplifying patch outliers.

We also use TokenRank to gate heads: at a candidate layer, we keep the top-$h$
heads by $\pi(\texttt{REG})$ (typically $h=1$--$3$) and apply the test-time register
only on those heads. This strengthens absorption and reduces image-token outliers,
yielding larger shifts in the $\lambda_2$--outlier joint plot than the vanilla baseline.
In practice, we find that the resulting absorption curves are qualitatively similar
across input images and classes, suggesting that the diagnostics capture a largely
backbone-specific rather than image-specific effect.

\subsection{Hyperparameter Sensitivity}
\label{secA:hyperparam_sensitivity}

We further evaluate the sensitivity of RegToken to key hyperparameters. As shown
in \Cref{tab:hyperparam_sensitivity}, performance remains stable over moderate
ranges of register-channel count, NFN-selected layers, outlier-token count,
initialization strength, and soft-bias strength. The best setting is not isolated
to a single brittle hyperparameter choice.

\begin{table}[t]
\centering
\caption{\textbf{Hyperparameter sensitivity.}
FID-5k under different RegToken hyperparameter settings. The method remains stable
across moderate ranges of channel count, NFN-selected layers, outlier-token count,
initialization strength, and soft-bias strength.}
\label{tab:hyperparam_sensitivity}
\small
\setlength{\tabcolsep}{4.0pt}
\renewcommand{\arraystretch}{1.05}
\resizebox{0.97\textwidth}{!}{
\begin{tabular}{lcc}
\toprule
\textbf{Hyperparameter} & \textbf{Values} & \textbf{FID-5k ($\downarrow$)} \\
\midrule
\# channels $k$ & 64 / 128 / 256 / 512 & 20.5 / 20.3 / \textbf{20.1} / 20.4 \\
NFN top-$\kappa$ layers & 1 / 3 / 5 & 20.4 / \textbf{20.1} / 20.2 \\
\# outlier tokens $q$ & 1 / 2 / 4 / 8 & 20.3 / \textbf{20.1} / 20.4 / 20.6 \\
Init strength $\gamma$ & 0.25 / 0.5 / 0.75 / 1.0 & 20.6 / 20.3 / \textbf{20.1} / 20.2 \\
Soft-bias $\beta$ & 0.5 / 1.0 / 2.0 & 20.4 / \textbf{20.1} / 20.3 \\
\bottomrule
\end{tabular}
}
\end{table}

\subsection{Fusion with \texttt{[CLS]}}
\label{secA:cls_fusion}

We also evaluate whether RegToken and \texttt{[CLS]} provide complementary global
signals. We form a fused prior
$u=\mathrm{norm}(\alpha u_{\mathrm{reg}} + (1-\alpha)u_{\mathrm{cls}})$ with
$\alpha\in\{0.25,0.5,0.75\}$. As shown in \Cref{tab:cls_fusion}, RegToken-heavy
fusion slightly improves over RegToken alone, while CLS-heavy fusion is less
stable. This indicates that RegToken supplies the main global prior and
\texttt{[CLS]} can add light semantic information.

\begin{table}[t]
\centering
\caption{\textbf{Fusion with \texttt{[CLS]}.}
Results are reported as DINOv2 / OpenCLIP. RegToken-heavy fusion performs best,
while CLS-heavy fusion is less stable.}
\label{tab:cls_fusion}
\small
\setlength{\tabcolsep}{4.0pt}
\renewcommand{\arraystretch}{1.05}
\resizebox{0.97\textwidth}{!}{
\begin{tabular}{lcccc}
\toprule
\textbf{Prior} & \textbf{FID-5k ($\downarrow$)} & \textbf{IS ($\uparrow$)} &
\textbf{CLIP ($\uparrow$)} & \textbf{SigLIP ($\uparrow$)} \\
\midrule
\texttt{[CLS]} & 20.5 / 20.8 & 281 / 283 & 0.39 / 0.40 & 3.6 / 3.5 \\
RegToken & 20.1 / 20.3 & 289 / 287 & 0.45 / 0.42 & 3.9 / 3.8 \\
RegToken + \texttt{[CLS]} ($\alpha=0.25$) & 20.4 / 20.7 & 282 / 283 & 0.41 / 0.40 & 3.5 / 3.6 \\
RegToken + \texttt{[CLS]} ($\alpha=0.50$) & 20.2 / 20.5 & 286 / 285 & 0.43 / 0.41 & 3.7 / 3.7 \\
RegToken + \texttt{[CLS]} ($\alpha=0.75$) & \textbf{20.0 / 20.2} & \textbf{289 / 288} & \textbf{0.45 / 0.43} & \textbf{4.0 / 3.8} \\
\bottomrule
\end{tabular}
}
\end{table}

\section{Component Summary and Practical Interpretation}
\label{secA:component_summary}

To complement the ablation results in \Cref{app:ablation}, we summarize the
practical role of each component in RegToken. Rather than introducing additional
experiments, this section distills the ablation trends into an interpretable view
of what each component contributes, which failure mode it addresses, and what
minimal configuration already provides a strong improvement over the baseline.

\subsection{What Each Component Contributes}
\label{secA:component_contrib}

\begin{table}[t]
\centering
\caption{\textbf{Practical Summary of the Main Components.}
A concise interpretation of the role of each component based on the ablations in
\Cref{tab:ablation_components,tab:ablation}.}
\label{tab:component_summary}
\small
\setlength{\tabcolsep}{5pt}
\renewcommand{\arraystretch}{1.08}
\resizebox{0.97\textwidth}{!}{
\begin{tabular}{p{3.0cm} p{4.7cm} p{4.4cm}}
\toprule
\textbf{Component} & \textbf{Main Contribution} & \textbf{Evidence} \\
\midrule
NFN-based layer selection &
Selects intervention layers where sink/register behavior is strongest, improving
where the prior is inserted. &
Adding NFN-based layer selection improves FID-5k from 21.5 to 21.3 in
\Cref{tab:ablation_components}; removing it degrades FID-5k to 21.1 in
\Cref{tab:ablation}. \\

TokenRank head gating &
Identifies heads that more effectively carry the injected prior and improves
prior routing. &
Adding TokenRank head gating further improves FID-5k from 21.3 to 20.8 in
\Cref{tab:ablation_components}; removing it degrades FID-5k to 20.9 in
\Cref{tab:ablation}. \\

Interpolation / conservation update &
Stabilizes the inserted prior representation and preserves useful global structure
during transfer to the decoder token space. &
Adding the interpolation step improves FID-5k from 20.8 to 20.3 and also slightly
improves IS and SigLIP in \Cref{tab:ablation_components}. \\

Register neuron selection ($k$) &
Provides a compact subspace for constructing the proposed prior while remaining
stable over a moderate range of selected dimensions. &
Varying $k$ from 16 to 64 changes FID-5k only mildly (20.7 to 20.3) in
\Cref{tab:ablation}. \\

Scale / strength ($s$ or $\gamma$) &
Controls the strength of prior insertion without requiring fine hyperparameter
tuning. &
Changing the scale from 1.0 to 4.0 only changes FID-5k from 20.9 to 21.0 in
\Cref{tab:ablation}. \\
\bottomrule
\end{tabular}
}
\end{table}

A concise practical summary of these component-wise contributions is provided in~\Cref{tab:component_summary}. The ablations show that the gain of RegToken is not tied to a single component.
Instead, each design choice contributes incrementally: NFN-based layer selection
improves intervention placement, TokenRank head gating improves head selection,
and the interpolation step yields the final performance gain by stabilizing the
inserted prior representation.

\subsection{Which Failure Mode Each Component Addresses}
\label{secA:component_failure_modes}

\begin{table}[t]
\centering
\caption{\textbf{Failure Modes Addressed by Each Component.}
Interpretation of the main failure mode mitigated by each design choice.}
\label{tab:failure_mode_summary}
\small
\setlength{\tabcolsep}{5pt}
\renewcommand{\arraystretch}{1.08}
\resizebox{0.97\textwidth}{!}{
\begin{tabular}{p{3.0cm} p{4.9cm} p{4.2cm}}
\toprule
\textbf{Component} & \textbf{Failure Mode Addressed} & \textbf{Practical Effect} \\
\midrule
NFN-based layer selection &
Inserting the prior at suboptimal layers where sink/register behavior is weak. &
Improves the consistency of intervention points across images. \\

TokenRank head gating &
Routing the prior through heads that do not effectively absorb or propagate the
inserted global token. &
Improves the usefulness of the inserted prior and reduces ineffective head usage. \\

Interpolation / conservation update &
Naively inserting the prior in a way that disrupts the decoder token space or fails
to preserve structured information. &
Makes prior transfer more stable and yields better final generation quality. \\

Register neuron selection &
Using an over-complete or under-complete set of feature dimensions when forming the
prior. &
Maintains a compact but robust prior representation. \\

Scale / strength &
Over-weak or over-strong prior injection that either has little effect or perturbs
decoding too aggressively. &
Keeps the method stable across a moderate range of insertion strengths. \\
\bottomrule
\end{tabular}
}
\end{table}

The corresponding mapping from each design choice to the failure mode it mitigates is summarized in~\Cref{tab:failure_mode_summary}. The ablation results suggest that RegToken works best when the prior is inserted
at the right layer, routed through informative heads, and transferred in a way
that preserves its structured global information.

\subsection{Minimal Working Configuration}
\label{secA:minimal_config}

A practical takeaway from~\Cref{tab:ablation_components,tab:ablation} is that
the full RegToken configuration provides the best overall performance, but a
simpler configuration already yields a meaningful gain over the baseline. In
particular, adding NFN-based layer selection and TokenRank head gating on top of
the TTR baseline reduces FID-5k from 21.5 to 20.8, even before applying the final
interpolation step. This suggests that a minimal working configuration can be
formed by combining NFN-based layer selection with TokenRank head gating, while
the interpolation / conservation update is the final ingredient needed to recover
the full gain of RegToken.

\section{Linear Probe on ImageNet}
\label{secA:linear_image}

Consistent with prior work on trained and test-time registers, register-based
features achieve competitive but slightly lower ImageNet linear-probe accuracy
than the \texttt{[CLS]} token, while clearly outperforming simple patch averaging.
This pattern suggests that registers encode compact scene-level statistics rather
than serving as the strongest discriminative readout. Accordingly, the main paper
focuses on repurposing these features as plug-and-play global priors for generation,
where global context such as style, lighting, and layout is more directly useful.
The quantitative comparison is summarized in \Cref{tab:linear_probe}.
\begin{table}[t]
\centering
\caption{\textbf{Linear-Probe Accuracy on ImageNet-1k (DINOv2-L/14).}
A linear classifier is trained on frozen features from different global summaries.
Results for trained registers and test-time registers are consistent with prior
reports~\cite{DarcetOMB24,abs-2506-08010}. The \texttt{CLS} token remains the
strongest discriminative readout, while register-based features retain non-trivial
semantic content.}
\label{tab:linear_probe}
\begin{tabular}{lc}
\toprule
\textbf{Feature} & \textbf{Top-1 Accuracy (\%) ($\uparrow$)} \\
\midrule
{[\texttt{CLS}] token}                          & 85.4 \\
Patch mean                                      & 84.4 \\
Trained register token~\cite{DarcetOMB24}       & 83.1 \\
Test-time register token~\cite{abs-2506-08010}  & 84.5 \\
RegToken (NFN + TokenRank register)             & 84.8 \\
\bottomrule
\end{tabular}
\end{table}
These results support the view that register-based summaries preserve meaningful
global information even when they are not the best classification readout, which
is consistent with their use as transferable priors for frozen tokenized generation.

\section{Failure Cases and Limitations} 
\label{secA:limit} 

\subsection{Failure Examples} 
\label{secA:failure_examples} 
\begin{figure}[t!] 
\centering 
\includegraphics[width=\textwidth]{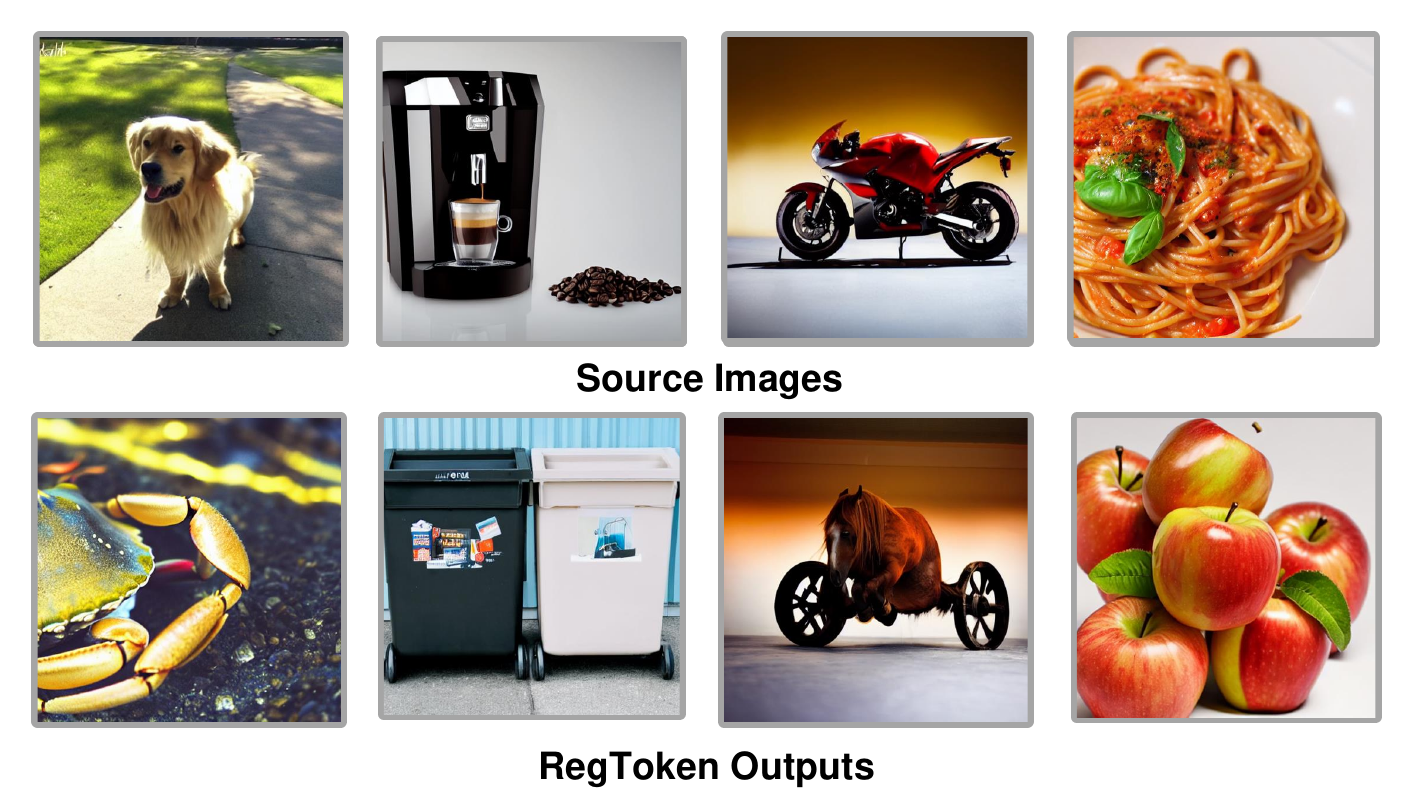} 
\caption{\textbf{Representative Failure Cases.} Top row: source images. Bottom row: corresponding RegToken outputs. The examples illustrate several characteristic failure modes, including incomplete target-object structure, source-to-target semantic drift, and category confusion under challenging source-target transfers. These cases are consistent with our analysis that RegToken mainly provides compact global structure rather than a full replacement for fine token-level modeling and precise geometric composition.} 
\label{fig:failure_cases} 
\end{figure}

We observe several failure cases in which the proposed prior does not fully resolve the limitations of the frozen decoding pipeline. Typical examples include strong source-induced semantic drift, target-category confusion, incomplete object structure, and failures under uncommon compositional or geometric demands. In such cases, the global prior may still shape coarse layout or appearance, but it does not reliably recover the exact target semantics or the full local part arrangement.~\Cref{fig:failure_cases} shows representative examples.

\subsection{Cases Where the Prior Helps Less}
\label{secA:helps_less}

The proposed prior is most useful when global structure plays an important role in
the decoded image, such as scene layout, illumination, or coarse semantic context.
Its benefit is less pronounced when the target generation depends primarily on
fine local texture, highly specific object instances, or precise geometric
arrangement. This is consistent with our analysis that register-based priors mainly
capture compact global statistics rather than serving as a complete replacement for
token-level optimization or richer generative modeling.

\subsection{Scope and Generalization Limits}
\label{secA:scope_limits}

This work focuses primarily on large vision transformers. Similar attention-sink
phenomena have been reported in large language models, multimodal models, and
retrieval architectures, but our method has not yet been evaluated in those settings.
In addition, while the proposed approach improves token-based generative pipelines,
more complex architectures with attention modules (\eg, DiT) may exhibit different
sink dynamics and may require different intervention strategies. Understanding how
register-like mechanisms interact with such models remains an important direction
for future work. At the same time, we believe the current scope is still practically meaningful: RegToken operates in a fully frozen setting, requires no retraining, and under the Prior + opt protocol updates only a single injected global token. Moreover, its benefit appears not only in final generation metrics but also in
optimization efficiency and stability, as reflected by Steps@$\tau$, AUC, and Std@S
in~\Cref{tab:opt_dynamics_summary}. We therefore view the present setting as a
useful testbed for lightweight plug-and-play priors in tokenized generation, while
leaving broader validation on other generative architectures to future work.

\section{Algorithms and Additional Generality Checks}
\label{secA:algor}

We summarize the three main procedures used in the paper:
\begin{itemize}
    \item \Cref{alg:nfn_dnr_tokenrank}: NFN-based layer and register-neuron localization with TokenRank diagnostics.
    \item \Cref{alg:reg2decode}: Register-guided decoding for 1D tokenizers such as HCT.
    \item \Cref{alg:valuetokenrank}: The Value-to-TokenRank bridge used to quantify register absorption and select heads for gating.
\end{itemize}

\begin{algorithm}[ht]
\small
\caption{NFN-based Localization and TokenRank.}
\label{alg:nfn_dnr_tokenrank}
\begin{algorithmic}[1]
\Require ViT, dataset $\mathcal{D}$, image batch $\mathcal{B}$, $\kappa,n,k,w,\Delta,s$
\State Compute $\mathrm{NFN}_{\ell,m}$ over $\mathcal{D}$ for each layer $\ell$ and module $m$
\State Aggregate $S_\ell=\max_m \mathrm{NFN}_{\ell,m}$ and pick
       $\mathcal{L}_{\mathrm{cand}}=\mathrm{top}\text{-}\kappa$ layers with largest $S_\ell$
\For{$\ell \in \mathcal{L}_{\mathrm{cand}}$} \Comment{Register-neuron extraction}
  \State Detect outlier tokens $\mathcal{P}$ via $\ell_2$-norm maps across a window of $w$ layers
  \State Rank channels by mean $|a_{p,d}|$ over $p\in\mathcal{P}$ and previous $\Delta$ layers
  \State Take $\mathcal{R}_\ell=\mathrm{top}\text{-}k$ channels as register neurons
  \State Form $t_{\mathrm{reg}}$ and update values by the projection-and-conservation update in Section~\ref{sec:interp} with scale $s$
\EndFor
\For{each head} \Comment{Value$\times$TokenRank diagnostics}
  \State Compute $\mathrm{WriteMass}$, TokenRank $\pi$, and $\lambda_2$ (second eigenvalue)
  \State Log curves of TokenRank(REG), WriteMass(REG), and $\lambda_2$ as $s$ varies
\EndFor
\State \Return Modified forward pass with test-time registers and per-head diagnostics
\end{algorithmic}
\end{algorithm}

\begin{algorithm}[ht]
\small
\caption{Register-guided Decoding for 1D tokenizers.}
\label{alg:reg2decode}
\begin{algorithmic}[1]
\Require Register tokens $\{r^{(\ell)}\}$, codebook $\mathcal{C}$, base tokens $\mathbf{z}$, weights $\{\alpha_\ell\}$, strength $\gamma$
\For{each $\ell\in\mathcal{L}_\mathrm{use}$}
  \State $j_\ell^\star \gets \arg\max_j\,\sigma(r^{(\ell)}, c_j)$,\quad $\tilde r^{(\ell)} \gets c_{j_\ell^\star}$
\EndFor
\State $\tilde r \gets \sum_{\ell}\alpha_\ell\,\tilde r^{(\ell)}$ \Comment{Fuse layer-wise registers}
\State $z_0 \gets (1-\gamma)z_0 + \gamma\,\tilde r$ \Comment{Insert as global prior token}
\State \Return decoded image $\mathcal{D}([z_0,\mathbf{z}_{1:T}])$
\end{algorithmic}
\end{algorithm}

\begin{algorithm}[ht]
\small
\caption{Value-to-TokenRank Bridge for Absorption.}
\label{alg:valuetokenrank}
\begin{algorithmic}[1]
\Require Attention matrices $\{A^{(\ell,h)}\}$, value tensors $\{V^{(\ell,h)}\}$,
         register index $t_{\mathrm{reg}}$, scales $\mathcal{S}=\{s_1,\dots,s_M\}$,
         number of gated heads $h_{\text{gate}}$
\For{each layer $\ell$ and head $h$}
  \For{each scale $s \in \mathcal{S}$}
    \State Scale register coordinates in $V^{(\ell,h)}$ by $s$ on the register subspace
    \State Recompute attention $A^{(\ell,h)}$ if needed and obtain updated values $V^{(\ell,h)}$
    \State Compute WriteMass($t_{\mathrm{reg}}$) via
           $\mathrm{WriteMass}(t)=\sum_i A^{(\ell,h)}_{i,t}\,\|V^{(\ell,h)}_t\|_2^2$
    \State Treat $A^{(\ell,h)}$ as a Markov chain; compute TokenRank $\pi^{(\ell,h)}$ (stationary dist.)
    \State Estimate the second eigenvalue $\lambda_2^{(\ell,h)}$ of $A^{(\ell,h)}$
    \State Log TokenRank($t_{\mathrm{reg}}$), WriteMass($t_{\mathrm{reg}}$), and $\lambda_2^{(\ell,h)}$ for scale $s$
  \EndFor
\EndFor
\State For each layer $\ell$, aggregate TokenRank($t_{\mathrm{reg}}$) over heads and
       select the top-$h_{\text{gate}}$ heads as gated heads
\State \Return Absorption curves (TokenRank/WriteMass/$\lambda_2$ vs.\ $s$) and head-gating sets
\end{algorithmic}
\end{algorithm}

\clearpage

\subsection{Generality Beyond the Main HCT-style Setting}
\label{secA:generality}

To complement the main HCT-style experiments, we evaluate whether the proposed prior generalizes beyond the primary decoder and backbone setting. We consider three axes: compact 1D token budgets based on TiTok/HCT-style tokenizers~\cite{YuWDSCC24,abs-2506-08257}, an alternative MaskGIT-VQGAN decoding interface~\cite{ChangZJLF22}, and DINOv2 backbone scales. The VQ-LL/VQ-BB/VQ-BL names denote the compact tokenizer variants used in this diagnostic check, and the suffix indicates the token budget.

\myparagraph{Compact 1D token budgets.}
We first test multiple compact-token variants with different token budgets. Each
cell reports results for DINOv2 / OpenCLIP priors under the same frozen decoding
protocol. Across 32-, 64-, and 128-token variants, RegToken consistently improves
over the corresponding no-prior and \texttt{[CLS]}-prior baselines.

\begin{table}[t]
\centering
\caption{\textbf{Generality across compact 1D token budgets.}
FID-5k results for different compact tokenizer variants. The VQ-LL/VQ-BB/VQ-BL names denote the tokenizer variants used in this diagnostic check, and the suffix indicates the token budget. Each cell reports DINOv2 / OpenCLIP. RegToken consistently improves over no-prior and \texttt{[CLS]}-prior baselines across token budgets.}
\label{tab:generality_token_budget}
\small
\setlength{\tabcolsep}{4.0pt}
\renewcommand{\arraystretch}{1.05}
\resizebox{0.97\textwidth}{!}{
\begin{tabular}{lcccc}
\toprule
\textbf{Tokenizer variant} & \textbf{No prior} & \textbf{\texttt{[CLS]} prior} &
\textbf{RegToken} & \textbf{$\Delta$FID over no prior} \\
\midrule
VQ-LL-32  & 21.2 / 21.7 & 20.5 / 20.8 & \textbf{20.1 / 20.3} & -1.1 / -1.4 \\
VQ-BB-64  & 21.8 / 22.3 & 21.3 / 21.7 & \textbf{20.9 / 21.1} & -0.9 / -1.2 \\
VQ-BL-64  & 22.5 / 23.0 & 22.0 / 22.4 & \textbf{21.7 / 21.9} & -0.8 / -1.1 \\
VQ-BL-128 & 23.5 / 24.0 & 23.0 / 23.6 & \textbf{22.6 / 23.3} & -0.9 / -0.7 \\
\bottomrule
\end{tabular}
}
\end{table}

\myparagraph{MaskGIT-VQGAN pilot.}
We also test an alternative MaskGIT-VQGAN~\cite{ChangZJLF22} decoding interface.
Since this pipeline does not use the same inserted HCT-style global slot, we map
the register-derived prior to a frozen codebook-logit initialization bias. As shown
in \Cref{tab:generality_maskgit}, RegToken improves over the MaskGIT baseline as
well as random and \texttt{[CLS]} initialization controls.

\begin{table}[t]
\centering
\caption{\textbf{MaskGIT-VQGAN pilot.}
RegToken is used as a frozen codebook-logit initialization bias. It improves FID,
CLIP, and SigLIP over MaskGIT, random-init, and \texttt{[CLS]}-init controls.}
\label{tab:generality_maskgit}
\small
\setlength{\tabcolsep}{4.0pt}
\renewcommand{\arraystretch}{1.05}
\resizebox{0.97\textwidth}{!}{
\begin{tabular}{lcccc}
\toprule
\textbf{Prior on MaskGIT} & \textbf{FID-1k ($\downarrow$)} & \textbf{CLIP ($\uparrow$)} &
\textbf{SigLIP ($\uparrow$)} & \textbf{($\Delta$) over no prior} \\
\midrule
MaskGIT baseline & 31.8 & 0.392 & 2.45 & -- \\
Random init bias & 31.6 & 0.393 & 2.46 & -0.2 / +0.001 / +0.01 \\
\texttt{[CLS]} init bias & 31.1 & 0.398 & 2.55 & -0.7 / +0.006 / +0.10 \\
RegToken init bias & \textbf{30.3} & \textbf{0.406} & \textbf{2.72} & -1.5 / +0.014 / +0.27 \\
\bottomrule
\end{tabular}
}
\end{table}

\myparagraph{Backbone scales.}
Finally, we evaluate RegToken across DINOv2 backbone scales. RegToken remains
stronger than the corresponding \texttt{[CLS]} prior across DINOv2-B/L/g, with
smaller gains on ViT-B and comparable gains on ViT-g.

\begin{table}[t]
\centering
\caption{\textbf{Generality across DINOv2 backbone scales.}
FID-5k comparison between \texttt{[CLS]} and RegToken priors across DINOv2
backbone sizes.}
\label{tab:generality_backbone}
\small
\setlength{\tabcolsep}{5pt}
\renewcommand{\arraystretch}{1.05}
\begin{tabular}{lccc}
\toprule
\textbf{Backbone on FID-5k ($\downarrow$)} & \textbf{\texttt{[CLS]}} &
\textbf{RegToken } & \textbf{Gain over \texttt{[CLS]}} \\
\midrule
DINOv2-L/14 & 20.5 & \textbf{20.1} & +0.4 \\
DINOv2-B/14 & 21.0 & \textbf{20.9} & +0.1 \\
DINOv2-g/14 & 20.3 & \textbf{19.9} & +0.4 \\
\bottomrule
\end{tabular}
\end{table}

Together, these results suggest that the benefit of RegToken is not limited to the
specific HCT-style decoder configuration used in the main experiments. Instead,
the proposed register-derived prior provides a reusable global signal across token
budgets, decoder interfaces, and backbone scales. Full diffusion or autoregressive
integration may require different injection interfaces and is left for future work.

% ---- Bibliography ----
%
% BibTeX users should specify bibliography style 'splncs04'.
% References will then be sorted and formatted in the correct style.
%

\end{document}

%% file: preamble.tex
\usepackage{url}
\usepackage{graphicx}
\usepackage{capt-of}
\usepackage{booktabs} 
\usepackage{multirow}
\usepackage{threeparttable }

\usepackage{makecell}
\usepackage{array}